\pdfoutput=1

\documentclass[10pt,twocolumn,letterpaper]{article}

\usepackage[pagenumbers]{cvpr} 

\usepackage{cuted}    
\usepackage{caption}
\usepackage{subcaption}
\usepackage{microtype}
\usepackage{array}
\usepackage{placeins}
\ifdefined\linenumbersep
  \leftlinenumbers*
\fi
\graphicspath{{fig/}}

\definecolor{cvprblue}{rgb}{0.21,0.49,0.74}
\usepackage[pagebackref,breaklinks,colorlinks,allcolors=cvprblue]{hyperref}

\def\confName{CVPR}
\def\confYear{2027}

\newcommand{\cmark}{\textcolor{ckgreen}{\checkmark}}
\newcommand{\xmark}{\textcolor{black!35}{--}}

\usepackage{colortbl}
\usepackage{tabularx}
\usepackage{multirow}
\definecolor{ckgreen}{HTML}{2F6F57}
\definecolor{bestcell}{HTML}{A9D3B8}
\definecolor{secondcell}{HTML}{D3EADB}
\definecolor{headgray}{HTML}{EEF1EF}
\definecolor{oursrow}{HTML}{E4F1E9}
\definecolor{humanrow}{HTML}{F5EDE3}
\definecolor{gaintext}{HTML}{2F6F57}
\definecolor{depcol}{HTML}{B5733A}
\definecolor{thirdcell}{HTML}{E3F2E8}
\newcommand{\best}[1]{\cellcolor{bestcell}\textbf{#1}}
\newcommand{\second}[1]{\cellcolor{secondcell}#1}
\newcommand{\third}[1]{\cellcolor{thirdcell}#1}
\newcommand{\gain}[1]{\,{\scriptsize\textcolor{gaintext}{#1}}}
\newcolumntype{Y}{>{\centering\arraybackslash}X}
\newcolumntype{L}{>{\raggedright\arraybackslash}X}

\title{CrossProjection: Benchmarking Spatial Grounding\\across Projection and Section Views}

\providecommand{\iflatexml}{\iffalse}
\iflatexml
\author{Kaho Li*\\Tsinghua University \and
Zekai Lin*\\University of Glasgow \and
Pengyu Zeng\\Tsinghua University \and
Yuqin Dai\\Tsinghua University \and
Jun Yin\\Tsinghua University \and
Tianjing Feng\\University College London, University of London \and
Shuai Lu\\Tsinghua University}
\else
\author{%
Kaho Li\textsuperscript{1,*}\quad
Zekai Lin\textsuperscript{2,*}\quad
Pengyu Zeng\textsuperscript{1}\quad
Yuqin Dai\textsuperscript{1}\quad
Jun Yin\textsuperscript{1}\quad
Tianjing Feng\textsuperscript{3}\quad
Shuai Lu\textsuperscript{1}\\[0.6em]
{\normalsize\textsuperscript{1}Tsinghua University\quad
\textsuperscript{2}University of Glasgow\quad
\textsuperscript{3}University College London, University of London}
}
\fi

\begin{document}
\maketitle
\newcommand{\crossprojectionteasercaption}{\textbf{CrossProjection overview.} Cross-view tasks, model performance, spatial diagnostics, and BIM-supervised learning. Radars pool Natural/BIM by answer format and subtype; the human reference uses Human300 (matched comparisons: \cref{sec:human}). Bars compare the accuracy of Qwen3.5-9B before and after BIM curriculum fine-tuning, on BIM and on Natural drawings not used for fine-tuning (\cref{sec:sft}).}
\iflatexml
\begin{center}\textit{*Equal contribution: Kaho Li and Zekai Lin.}\end{center}
\begin{figure*}
  \centering
  \includegraphics[width=\textwidth]{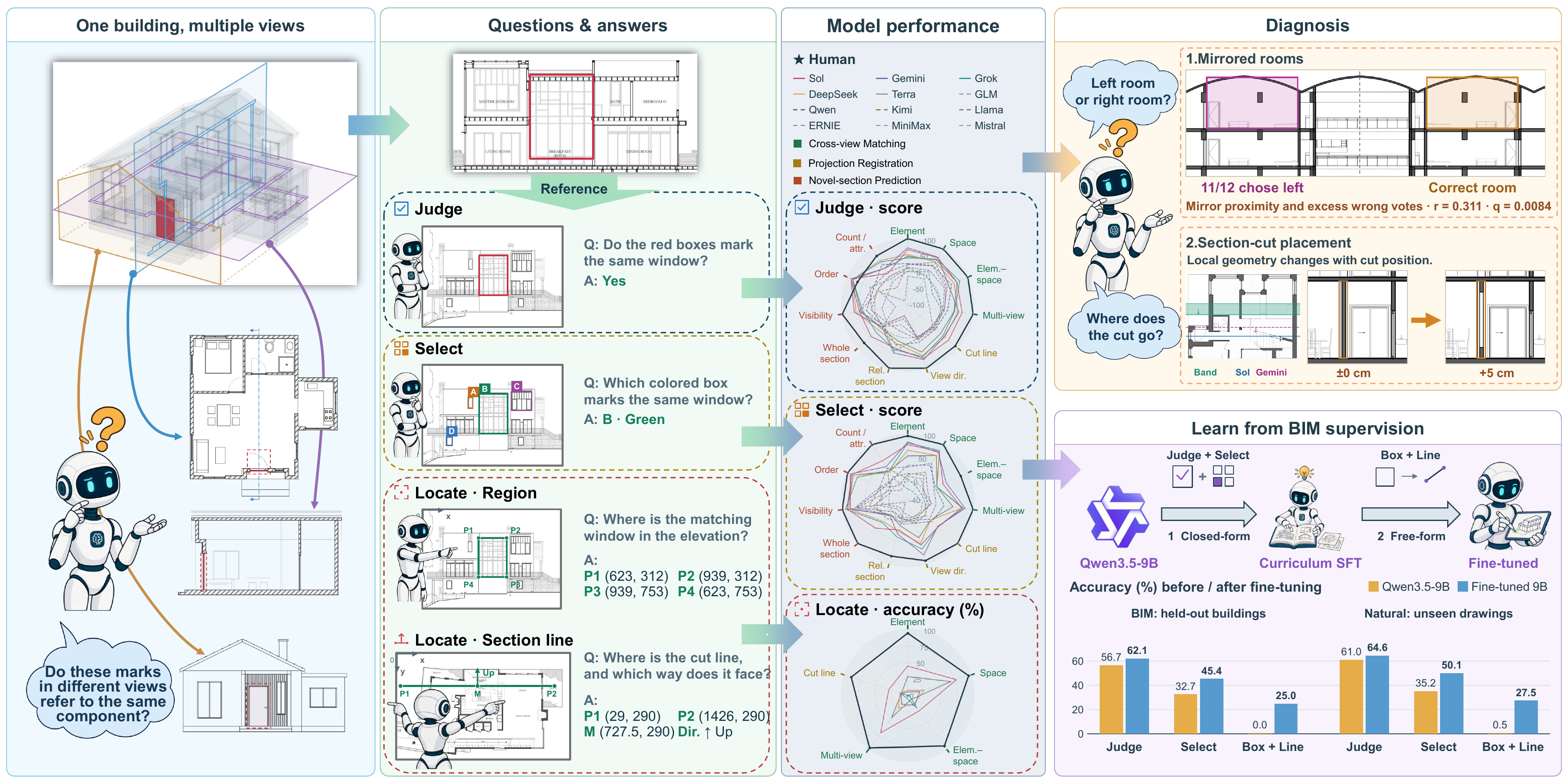}
  \caption{\crossprojectionteasercaption}
  \label{fig:teaser}
\end{figure*}
\else
{\let\thefootnote\relax\footnotetext{\textsuperscript{*}Equal contribution.}}

\begin{strip}
  \centering
  \includegraphics[width=\textwidth]{figure_01.pdf}
  \captionsetup{hypcap=false}
  \captionof{figure}{\crossprojectionteasercaption}
  \label{fig:teaser}
  \captionsetup{hypcap=true}
\end{strip}
\fi

\begin{abstract}
Plans, elevations, and sections describe a three-dimensional structure through incomplete projections that change how an entity is drawn, not only where it is seen from. Reading them means deciding which entities correspond across views and recovering where they lie. We introduce \textbf{CrossProjection}, a benchmark that uses architectural drawings to test whether vision-language models (VLMs) can turn correspondence decisions into geometric answers. Its tasks cover cross-view matching, projection registration, and novel-section prediction, with outputs ranging from judgments and candidate selection to free region boxes and directed section cuts. For inverse section localization, we introduce, to our knowledge, the first VLM benchmark protocol that accepts any cut within an expert-adjudicated band of compatible positions and checks the viewing direction separately. Manually annotated drawings and BIM-derived views provide 33,788 questions from 46 buildings, and we evaluate 22 VLMs on a common set of 2,763. Strong selection or reproduction does not imply cross-view localization: on 100 matched region targets, the best model redraws 98 marked boxes but localizes only 59 across views. Section cuts remain far from expert level, with the leading model trailing two architects by 51.9 and 67.4 percentage points on shared Natural and BIM line questions. Incorrect choices favor mirrored positions, and many cuts face the correct way but violate local geometry. Curriculum fine-tuning on BIM supervision teaches a 9B model to localize regions it could not localize before, raising box accuracy from 0\% to 25.3\% on held-out BIM buildings and from 0.3\% to 30.8\% on real drawing sets not used for fine-tuning, whereas section cuts remain largely out of reach.
\end{abstract}

\section{Introduction}
\label{sec:intro}

Three-dimensional structures are often communicated through several incomplete views. An orthographic projection discards depth along its viewing direction, and a section shows what a cutting plane intersects and what lies beyond it~\cite{ching2015architectural}. Reading such views together means working out which objects or spaces correspond across them and where each counterpart lies. Unlike camera images, projections and sections can depict the same entity with different graphical conventions, so appearance alone often cannot settle correspondence. A model may therefore judge that two marks refer to the same entity, or pick the matching candidate, without being able to recover where that entity lies in another projection or which cut produced a given section (\cref{fig:teaser}). The distinction matters in practice: as VLMs are asked to read and act on design documents, for example in drawing-guided CAD and BIM workflows, they must return geometric answers rather than choose among supplied options. We ask how far the correspondence decisions of current VLMs translate into geometrically consistent answers.

Existing benchmarks probe spatial relations and cross-view correspondence mainly in camera images~\cite{fu2024blink,yang2025thinking,yeh2025seeing,yang2025mmsi,wang2025crosspoint}. Benchmarks on technical drawings test drawing reading~\cite{kondratenko2026aecv,jung2026drawingvqa}, building identity across media~\cite{parikh2026archb}, candidate matching~\cite{shen2026archsibench}, and projection or section reasoning on solids and parts~\cite{wang2025spatialviz,niu2025creftcad}. Few of them require the answer itself to be geometric. CrossProjection goes beyond judgment and selection to free region localization and the recovery of directed cuts in complex interiors (\cref{tab:comparison}).

Architectural drawings make a natural testbed. They are among the most widely produced orthographic documents, and a single drawing set describes one building through plans, elevations, and sections that share repeated components, multiple storeys, and dense interiors. Because projection and intersection relations are explicit, any proposed correspondence or cut can be checked against the drawings. CrossProjection covers cross-view matching, projection registration, and novel-section prediction on two complementary sources. \emph{Natural} questions keep real drafting conventions, with manually delineated regions and cross-view instance links that make every target traceable to human annotation. \emph{BIM} questions are derived from building models that retain entity identity and geometry, which makes supervision scalable. In total, the resource contains 33,788 questions from 46 buildings, of which 2,763 form the common evaluation set.

The tasks are organized by how much geometry the answer must contain. Judge assesses a stated relation, Select picks the corresponding candidate, and region Locate must place the target box without any candidates. When neighboring or mirrored instances are present, Locate therefore requires recovering both the correct instance and its extent. Section localization goes one step further. Recovering a directed cut from a reference section requires a position and a viewing side that together reproduce the intersected and visible content, and several positions can be equally compatible. We therefore introduce \emph{accepted positional bands}, adjudicated by experts from local geometry, which accept any cut within the compatible range while checking the viewing direction separately. To our knowledge, this is the first VLM benchmark protocol for section localization built on such bands. Together with forward prediction of section content, it probes section geometry from both directions.

Our results follow this progression. Strong candidate selection does not ensure free localization, even for identical targets: on 134 matched Natural targets, Gemini 3.1 Pro and Grok-4.6 select the correct candidate for 120 and 119 but localize only 16 and 0. Nor does reproducing a marked region ensure finding it in another view. GPT-5.6 Sol (hereafter Sol), the strongest region localizer, redraws 100\% of marked boxes on Natural and 96\% on BIM in 100 matched controls (50 per source), yet localizes only 66\% and 52\% of the same targets across views. Section cuts are harder still. The best line accuracies among the 22 evaluated models are 40.0\% and 17.5\%, and on the shared line questions of the 300-question Human300 sample, Sol reaches 38.5\% and 21.7\% while two architects average 90.4\% and 89.1\%.

Spatial diagnostics show what these failures look like: incorrect choices drift toward mirrored positions, and many cuts face the correct way but violate local geometry. CrossProjection also serves as supervision. Curriculum fine-tuning on BIM data teaches a 9B model to localize regions that it could not localize before, both in held-out BIM buildings and in real drawing sets not used for fine-tuning. Our contributions are:
\begin{itemize}
  \item \textbf{A benchmark connecting correspondence and geometry}, using architectural projections to evaluate Judge and Select, free region localization, and directed section-cut localization within one task framework.
  \item \textbf{A protocol for inverse section localization} that evaluates predicted cut positions against expert-adjudicated bands of geometrically compatible positions and checks viewing direction independently.
  \item \textbf{An evaluation of 22 VLMs and a diagnosis of 12 frontier models}, combining matched single-image controls, responses from two architects on 300 shared questions, and analyses of spatially structured errors.
  \item \textbf{Manually annotated cross-view data and reusable construction workflows}, combining manually established instance correspondences in native drawings with BIM questions derived from geometry, and a fine-tuning study showing that BIM supervision teaches region localization that transfers to real drawings.
\end{itemize}

\begin{table*}[t]
  \centering
  \footnotesize
  \setlength{\tabcolsep}{3.2pt}
  \setlength{\aboverulesep}{0pt}\setlength{\belowrulesep}{0pt}
  \renewcommand{\arraystretch}{1.15}
  \caption{Task coverage and ground-truth (GT) construction across benchmarks (Supplement A.6). \cmark/\xmark: documented/not documented; Human: original GT provided by people; \textcolor{depcol}{Dep.}: generative models take part in building reference answers or target annotations; ``?'': unclear from the paper.}
  \label{tab:comparison}
  \begin{tabular}{@{}llcccccccc@{}}
    \toprule
    \rowcolor{headgray}& & Multi-image / & Projection / & Judge / & Free & Localization & \multicolumn{3}{c@{}}{GT construction} \\
    \cmidrule(l){8-10}
    \rowcolor{headgray}Benchmark & Domain & 3D input & section & choice & localization & output & Human & Geometric & No gen.\ model \\
    \midrule
    AECV-Bench~\cite{kondratenko2026aecv} & Architecture & \xmark & \xmark & \xmark & \xmark & \xmark & \cmark & \xmark & \textcolor{depcol}{Dep.} \\
    ArchSIBench~\cite{shen2026archsibench} & Architecture & \cmark & \cmark & \cmark & \xmark & \xmark & \cmark & \xmark & \cmark \\
    ARCH-B~\cite{parikh2026archb} & Architecture & \cmark & \xmark & \cmark & \xmark & \xmark & \cmark & \xmark & \cmark \\
    DrawingVQA~\cite{jung2026drawingvqa} & Architecture & \cmark & \cmark & \cmark & \xmark & \xmark & \cmark & \xmark & \cmark \\
    \midrule
    LED-BENCH~\cite{li2026ledbench} & CAD / geometry & \cmark & \xmark & \xmark & \xmark & \xmark & \xmark & \xmark & \textcolor{depcol}{Dep.} \\
    MechVQA~\cite{kou2026mechvqa} & CAD / geometry & \cmark & \cmark & \xmark & \xmark & \xmark & \cmark & \xmark & \textcolor{depcol}{Dep.} \\
    TriView2CAD~\cite{niu2025creftcad} & CAD / geometry & \cmark & \cmark & \xmark & \xmark & \xmark & \xmark & \cmark & \textcolor{black!45}{?} \\
    SpatialViz-Bench~\cite{wang2025spatialviz} & CAD / geometry & \cmark & \cmark & \cmark & \xmark & \xmark & \cmark & \cmark & \cmark \\
    \midrule
    MedSG-Bench~\cite{yue2025medsg} & Medical & \cmark & \xmark & \xmark & \cmark & Box & \cmark & \cmark & \textcolor{black!45}{?} \\
    SPAR-Bench~\cite{zhang2025flatland} & Scenes & \cmark & \xmark & \cmark & \xmark & \xmark & \xmark & \cmark & \textcolor{black!45}{?} \\
    CrossPoint-Bench~\cite{wang2025crosspoint} & Scenes & \cmark & \xmark & \cmark & \cmark & Point & \xmark & \cmark & \textcolor{depcol}{Dep.} \\
    \midrule
    \rowcolor{oursrow}\textbf{CrossProjection (ours)} & \textbf{Architecture} & \cmark & \cmark & \cmark & \cmark & \textbf{Box + line} & \cmark & \cmark & \cmark \\
    \bottomrule
  \end{tabular}
\end{table*}

\ifdefined\FigureTwoSingleColumn
\begin{figure}[t]
\else
\begin{figure*}[t]
\fi
  \centering
  \ifdefined\FigureTwoSingleColumn
  \includegraphics[width=\columnwidth]{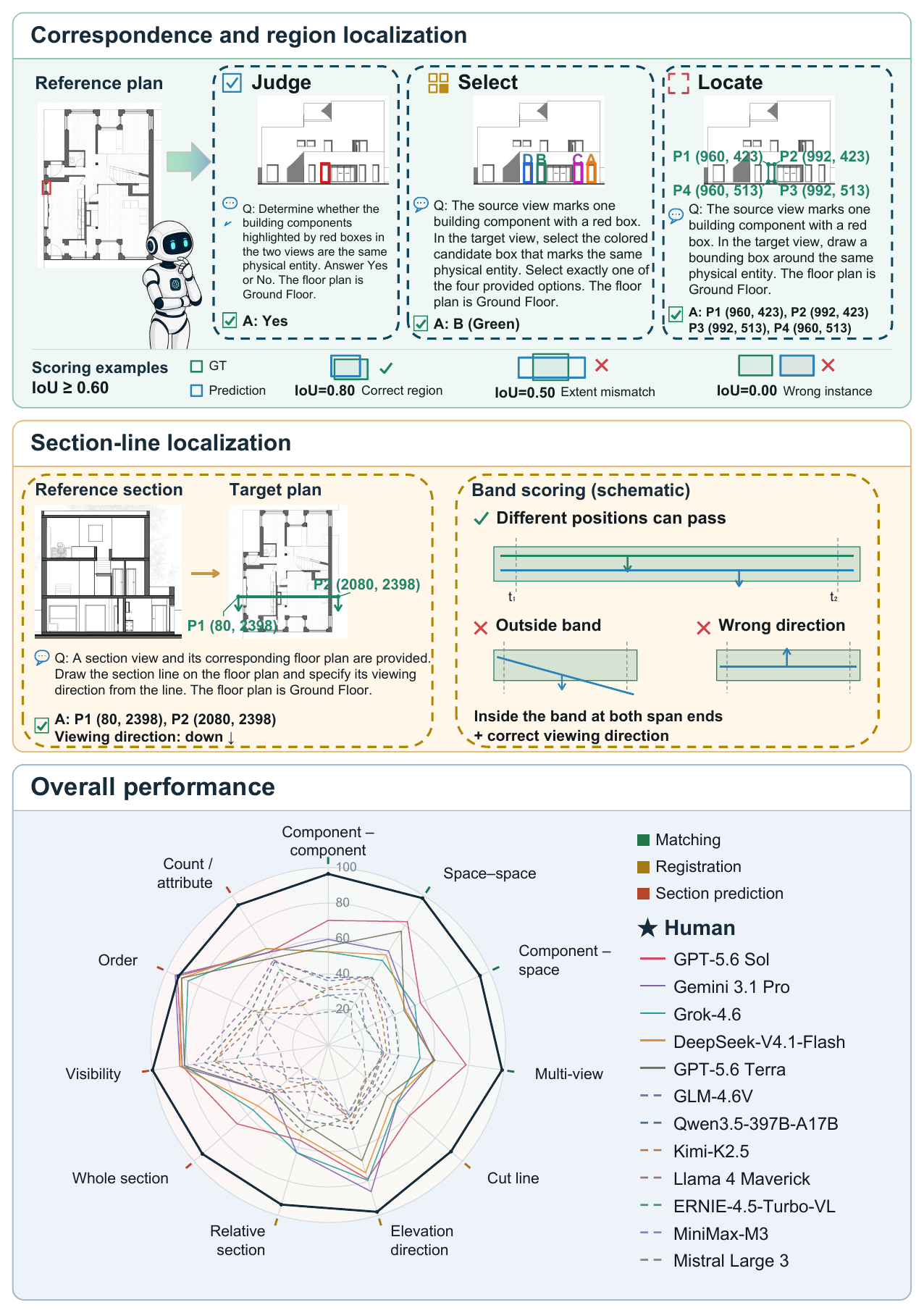}
  \else
  \includegraphics[width=\textwidth]{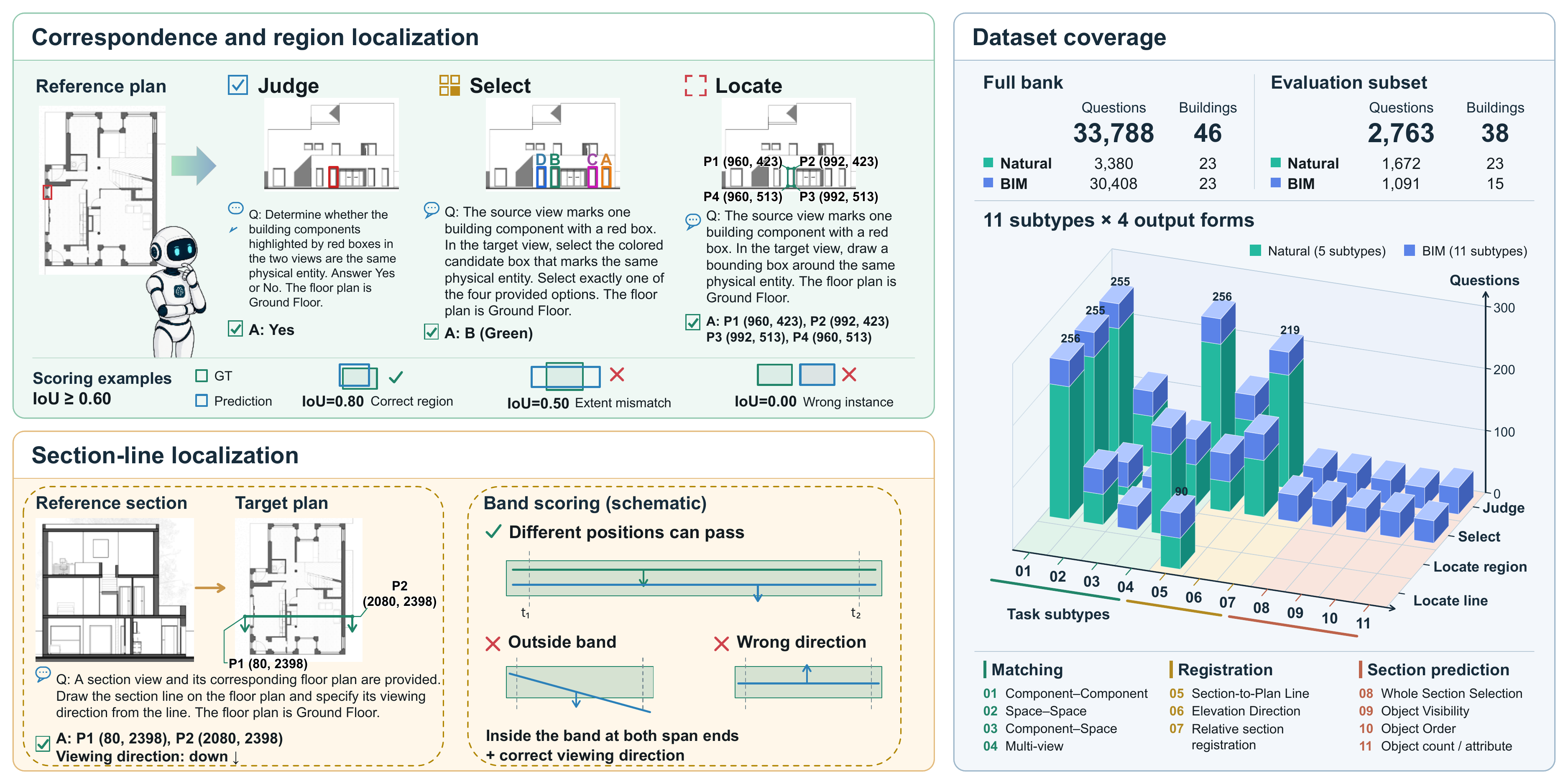}
  \fi
  \caption{Task examples, geometric scoring, and dataset coverage. \ifdefined\FigureTwoSingleColumn Top:\else Left:\fi\ correspondence judgment and selection, region localization, and section-line localization; coordinates refer to uncropped images. \ifdefined\FigureTwoSingleColumn Bottom:\else Right:\fi\ sizes of the full question bank and the evaluation subset, and evaluation questions per subtype and output form. Bar height gives the number of questions, with Natural (teal) and BIM (blue) stacked. Of the 44 combinations, 27 are supported; empty cells mark unsupported ones. Building counts include drawing sets.}
  \label{fig:overview}
\ifdefined\FigureTwoSingleColumn
\end{figure}
\else
\end{figure*}
\fi

\section{Related Work}
\label{sec:related}

\paragraph{Spatial reasoning in VLMs.}
A growing body of work probes and improves the spatial abilities of VLMs. SpatialVLM~\cite{chen2024spatialvlm} and SpatialRGPT~\cite{cheng2024spatialrgpt} train models on spatial question answering derived from 3D-lifted images, while BLINK~\cite{fu2024blink}, CV-Bench~\cite{tong2024cambrian}, and Eyes Wide Shut~\cite{tong2024eyes} show that even strong multimodal models miss basic visual and spatial cues. Later benchmarks extend the evaluation to 3D relations in natural images (3DSRBench~\cite{ma2025threedsrbench}), scenes and videos (VSI-Bench~\cite{yang2025thinking}, SPAR~\cite{zhang2025flatland}), and multiple views, including perspective taking, mental modeling, and cross-view point correspondence (All-Angles Bench~\cite{yeh2025seeing}, MMSI-Bench~\cite{yang2025mmsi}, ViewSpatial-Bench~\cite{li2025viewspatial}, MindCube~\cite{wang2025mindcube}, CrossPoint-Bench~\cite{wang2025crosspoint}). All of them operate on camera imagery. CrossPoint-Bench~\cite{wang2025crosspoint} already reveals a gap between discrete correspondence judgments and continuous point prediction. CrossProjection studies this gap in orthographic projections and sections, where correspondence must be recovered from geometry and drafting conventions, and extends the answers to region extents and directed cuts with multiple compatible positions.

\paragraph{Architectural and technical drawings.}
Computer vision has long analyzed technical drawings, recovering walls, rooms, and symbols from single floor plans~\cite{liu2017raster,kalervo2019cubicasa5k} and reconstructing 3D shapes from three orthographic views~\cite{hu2023plankassembly}. With the rise of VLMs, the focus has shifted to drawing reading and question answering (AECV-Bench~\cite{kondratenko2026aecv}, DrawingVQA~\cite{jung2026drawingvqa}, LED-BENCH~\cite{li2026ledbench}, MechVQA~\cite{kou2026mechvqa}). ArchSIBench~\cite{shen2026archsibench} evaluates architectural spatial intelligence, including correspondence through candidate selection, and ARCH-B~\cite{parikh2026archb} studies building identity from whole images. SpatialViz-Bench~\cite{wang2025spatialviz} tests cross-sections of compound solids, and CReFT-CAD with its TriView2CAD benchmark~\cite{niu2025creftcad} targets orthographic dimensions and parameters. CrossProjection complements these efforts by requiring free-form geometric grounding across projections and sections of complex interiors, together with forward prediction of section content and inverse cut localization.

\paragraph{Grounding and geometric answers.}
Grounded VLMs can output boxes or points in an image~\cite{peng2023kosmos2,you2024ferret,bai2025qwen25vl,deitke2025molmo}, and LocVLM~\cite{ranasinghe2024locvlm} shows that coordinate supervision also improves spatial reasoning. Beyond single images, MedSG-Bench~\cite{yue2025medsg} evaluates region grounding across images, and BOP-Distrib~\cite{meden2026bopdistrib} handles pose ambiguity that depends on the observation by representing the ground truth as a distribution. Inverse section localization faces an analogous ambiguity, since many cut positions can be compatible with one section. We represent this set as an accepted positional band and require both an in-band cut position and a correct viewing direction.

\section{The CrossProjection Benchmark}
\label{sec:benchmark}

\subsection{Tasks and Answer Formats}
\label{sec:tasks}

We organize the tasks around the five basic spatial abilities of Xu \etal~\cite{xu2025basic}: spatial perception, spatial relation, spatial orientation, mental rotation, and spatial visualization. In drawings, these abilities surface as three kinds of problems, which form our task categories. \emph{Cross-view Matching} asks whether a model can identify the same entity, and its spatial relations, across views. \emph{Projection Registration} asks whether it can carry positions and directions from one view to another. \emph{Novel-section Prediction} asks whether it can infer the structure and visible content produced by a specified cut.

Each category is refined into subtypes according to what must be matched, registered, or predicted, giving 11 in total (\cref{fig:overview}). Cross-view Matching has four (component--component, space--space, component--space, and multi-view correspondence), Projection Registration has three (section-to-plan line, elevation direction, and relative section registration), and Novel-section Prediction has four (whole-section selection, object visibility, object order, and object count or attribute).

A question supplies drawings, a reference or marked target, textual conditions, and optional candidates, and asks for an identity, relation, region, or section position (full definitions in Supplement A). The same spatial query can be answered at three levels of output demand (\cref{fig:overview}). \textbf{Judge} answers a relation or category, for instance whether two marked objects are the same physical instance. \textbf{Select} chooses the corresponding candidate. \textbf{Locate} must produce the answer itself, as a region box or a section line with viewing direction. In every format, repeated objects force identity to be resolved through positions and relations, and sections further require the cut position, viewing side, and depicted content to agree.

\subsection{Data Construction and Quality}
\label{sec:data}

\begin{figure*}[t]
  \centering
  \includegraphics[width=\textwidth]{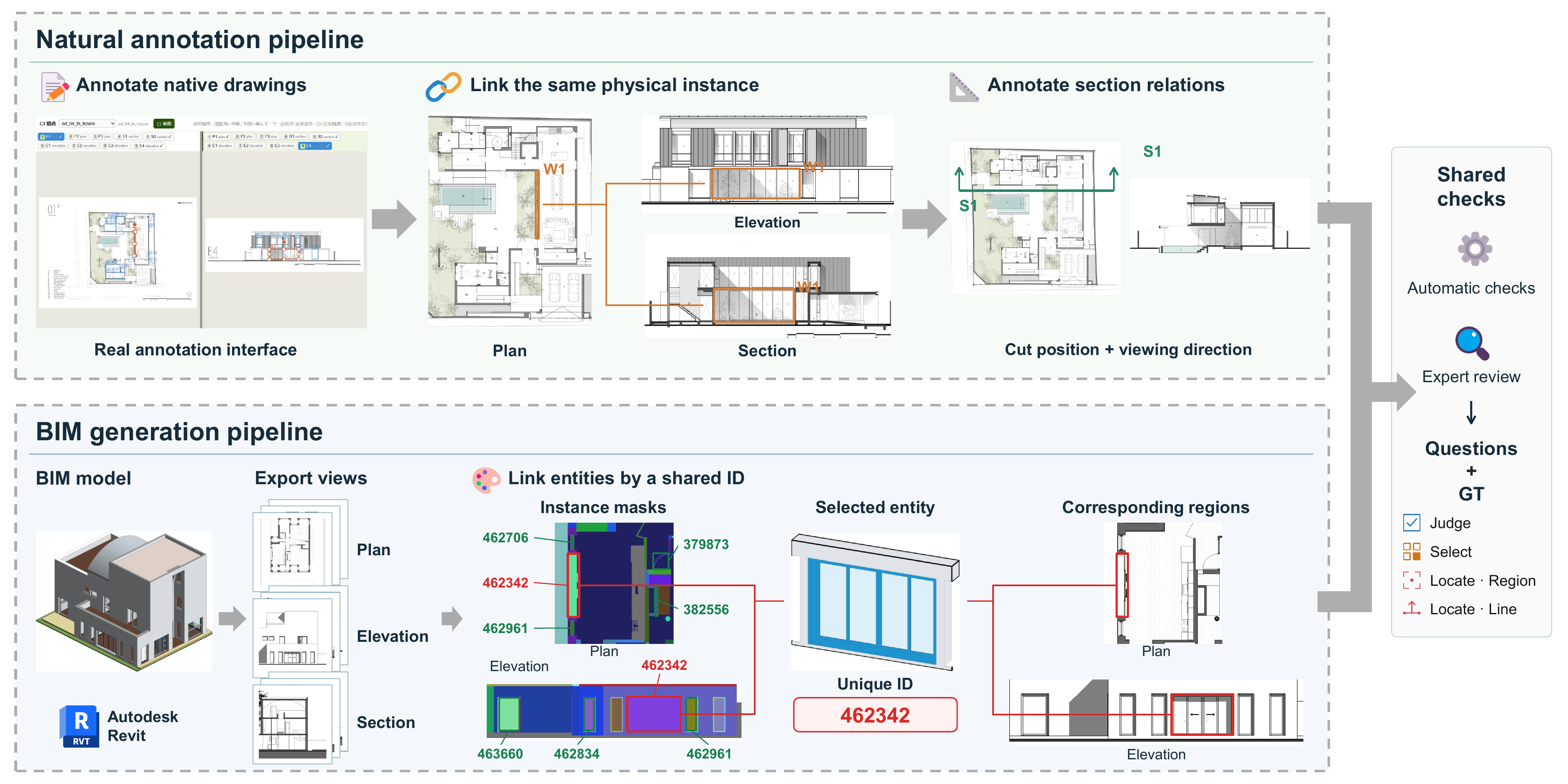}
  \caption{Human annotation and BIM question-generation pipelines. Natural manually links instances across drawings and annotates section relations; BIM uses instance masks and unique IDs to associate visible regions across views. In the BIM pipeline, red marks the selected entity and green other indexed entities. Both pipelines produce questions and GT through automatic checks and expert review.}
  \label{fig:construction}
\end{figure*}

CrossProjection draws on two complementary sources. \textbf{Natural} questions are constructed from human annotations on real drawing sets from ArchDaily, FreeFarmhouse, and YourHome. Annotated regions, cross-view anchors, and section relations explicitly link physical instances across drawings while preserving native drafting conventions. These annotations support 3,380 questions, including 1,672 of the 2,763 evaluation questions (60.5\%). \textbf{BIM} questions trade this diversity for scale, with 30,408 questions from 23 buildings. Their views are rendered from building models that retain entity identity, projections, cutting geometry, and visible pixels, so answers can be derived from the model itself, and human verification checks that the public drawings actually support them. Across both sources, ground truth is grounded in human annotation and deterministic geometry, without reference answers from generative models. The two pipelines share verification to produce evaluation and supervision data with traceable GT (\cref{fig:construction}; Supplement B).

We sample 2,763 questions from 38 buildings into a common evaluation set that covers both sources, all task subtypes, and all answer formats (\cref{fig:overview}). It contains 1,672 Natural questions from 23 drawing sets, covering five subtypes, and 1,091 BIM questions from 15 buildings, covering all eleven. Quality control combines automated consistency checks, an in-depth review of 300 questions by two experts, and a stratified check of 100 further questions, which found no substantive issue. Problems identified in the in-depth review were corrected throughout the bank.

\subsection{Accepted Geometry and Scoring}
\label{sec:scoring}

Judge and Select are scored by matching the predicted answer, whereas Locate requires the predicted geometry to match the target's position and extent. In every case, strict correctness demands both a valid output and a passing answer. Region boxes pass at IoU $\geq 0.60$, a threshold the annotated extents support: on Human300 (\cref{sec:setup}), the two architects reach it on 97.8\% of Natural and 93.8\% of BIM box questions.

Section lines need more care, because a single section can be compatible with several cut positions. Experts therefore infer from the plan what a candidate cut would reveal, compare it with the reference section, and mark the range of compatible positions, which forms a single contiguous band in every question, with two boundaries, an evaluation span, and a viewing direction. A predicted line passes when its direction is correct and its supporting line stays within the band, boundaries included, at both ends of the span, which tolerates the slight tilt of a hand-placed line. Two experts annotated all 90 line questions independently; their bands have median interval IoUs of 0.90 and 0.89 on Natural and BIM, and give identical pass/fail decisions for 99.5\% of 1,080 model responses. Primary scoring uses the first expert's bands. Model rankings are also stable under the scoring choices: Sol leads box localization on both sources at IoU thresholds of 0.50, 0.60, and 0.70, and widening every band by one pixel changes no model verdict (Supplement D.4).

Results are reported separately by source and answer format. Because the formats have different chance levels, raw strict accuracy $A$ on Judge and Select is converted to the chance-corrected score $S=100(A-b)/(1-b)$, with $b=0.50$ for binary judgments and $b=0.25$ for selection among four options. Box and line localization report accuracy (\%). Some diagnostics also use a \emph{content criterion} that first applies predefined deterministic format recovery.

\section{Experiments}
\label{sec:evaluation}

\subsection{Setup}
\label{sec:setup}

We evaluate 22 VLMs on the same 2,763 questions (\cref{tab:results}). Twelve are frontier proprietary and open-weight models, used with high reasoning effort where supported and default settings otherwise; in the text we refer to them by short names, \eg, Terra for GPT-5.6 Terra. The other ten are smaller open-weight models from 4B to 24B parameters, run locally with their native processors: Qwen3.5-9B~\cite{qwen35}, Qwen3-VL-4B/8B~\cite{bai2025qwen3vl}, Qwen2.5-VL-7B~\cite{bai2025qwen25vl}, InternVL3.5-8B/14B~\cite{wang2025internvl35}, Molmo2-4B/O-7B~\cite{clark2026molmo2}, Pixtral-12B~\cite{agrawal2024pixtral}, and Mistral Small 3.2~\cite{mistral2025small32} (settings in Supplement C). The diagnostics in \cref{sec:controls,sec:analysis} use the 12 frontier models. As an expert reference, two professional architects independently answer the same 300 questions (Human300); neither took part in question review or answer adjudication.

\begin{table*}[t]
  \centering
  \small
  \setlength{\tabcolsep}{4pt}
  \setlength{\aboverulesep}{0pt}\setlength{\belowrulesep}{0pt}
  \renewcommand{\arraystretch}{1.0}
  \caption{Main results of 22 VLMs on the 2,763 evaluation questions. Judge/Select: chance-corrected scores (points; 0 = random, 100 = perfect; negative values fall below chance, mostly because outputs that violate the answer format count as incorrect). Box/Line: strict accuracy (\%). $^\dagger$Proprietary; all other models are open-weight. Green shading marks the top three models per column within each block, darker for higher ranks (ties share a rank); the smaller models localize at most one target per column, so their Box/Line columns are not ranked. The last two rows compare Sol with the mean of two architects on the Human300 sample.}
  \label{tab:results}
  \fontsize{7}{10}\selectfont
  \begin{tabularx}{\textwidth}{@{}>{\raggedright\arraybackslash}p{0.205\textwidth} YYYY YYYY@{}}
    \toprule
    \rowcolor{headgray} & \multicolumn{4}{c}{\textbf{Natural}} & \multicolumn{4}{c@{}}{\textbf{BIM}} \\
    \rowcolor{headgray} & \multicolumn{2}{c}{\fontsize{6}{10}\selectfont  chance-corrected (pts)} & \multicolumn{2}{c}{\fontsize{6}{10}\selectfont  accuracy (\%)} & \multicolumn{2}{c}{\fontsize{6}{10}\selectfont  chance-corrected (pts)} & \multicolumn{2}{c@{}}{\fontsize{6}{10}\selectfont  accuracy (\%)} \\
    \rowcolor{headgray} Model & Judge & Select & Box & Line & Judge & Select & Box & Line \\
    \midrule
    \rowcolor{white} \multicolumn{9}{@{}l}{\textit{Frontier models}} \\
    \rowcolor{white} GPT-5.6 Terra$^\dagger$ & 52.9 & 56.5 & \second{40.2} & \third{16.0} & 28.3 & 39.1 & \second{31.3} & \third{2.5} \\
    \rowcolor{black!6} GPT-5.6 Sol$^\dagger$ & \best{73.8} & \third{74.6} & \best{57.8} & \best{40.0} & \best{43.8} & \best{58.5} & \best{54.0} & \best{17.5} \\
    \rowcolor{white} Gemini 3.1 Pro$^\dagger$ & \second{70.3} & \best{83.1} & 9.9 & \third{16.0} & \second{42.0} & \second{57.3} & \third{16.6} & \third{2.5} \\
    \rowcolor{black!6} Grok-4.6$^\dagger$ & 55.2 & \second{79.2} & 2.5 & \third{16.0} & \third{41.0} & \third{55.6} & 5.5 & \second{5.0} \\
    \rowcolor{white} DeepSeek-V4.1-Flash & \third{59.9} & 64.8 & \third{16.5} & \second{22.0} & 39.2 & 49.3 & 15.3 & 0.0 \\
    \rowcolor{black!6} Qwen3.5-397B-A17B & 23.9 & 32.0 & 1.5 & 0.0 & 22.9 & 23.3 & 0.6 & 0.0 \\
    \rowcolor{white} GLM-4.6V & 19.1 & 23.7 & 0.0 & 0.0 & 4.8 & 26.6 & 0.0 & 0.0 \\
    \rowcolor{black!6} Kimi-K2.5 & 20.4 & 13.0 & 0.8 & 2.0 & $-$2.5 & 10.8 & 0.0 & 0.0 \\
    \rowcolor{white} ERNIE-4.5-Turbo-VL & 8.6 & 3.1 & 0.3 & 2.0 & 6.6 & 9.0 & 0.0 & 0.0 \\
    \rowcolor{black!6} Mistral Large 3 & $-$13.9 & $-$9.1 & 0.0 & 0.0 & $-$23.4 & $-$5.6 & 0.0 & 0.0 \\
    \rowcolor{white} Llama 4 Maverick & 16.9 & 13.6 & 0.0 & 2.0 & $-$3.9 & 10.8 & 0.0 & 0.0 \\
    \rowcolor{black!6} MiniMax-M3 & 4.5 & 8.7 & 0.5 & 0.0 & $-$14.7 & 2.2 & 0.6 & 0.0 \\
    \midrule
    \rowcolor{white} \multicolumn{9}{@{}l}{\textit{Smaller open-weight models (4B--24B)}} \\
    \rowcolor{white} Qwen3.5-9B & \best{21.9} & \best{13.6} & 0.3 & 2.0 & \best{13.4} & 10.2 & 0.0 & 0.0 \\
    \rowcolor{black!6} Qwen3-VL-4B & \third{13.4} & \second{10.2} & 0.0 & 0.0 & \best{13.4} & 8.1 & 0.6 & 0.0 \\
    \rowcolor{white} Qwen3-VL-8B & \third{13.4} & \third{8.0} & 0.0 & 0.0 & 9.8 & 5.1 & 0.0 & 0.0 \\
    \rowcolor{black!6} Qwen2.5-VL-7B & \second{15.1} & 6.5 & 0.0 & 2.0 & 1.1 & 5.4 & 0.6 & 0.0 \\
    \rowcolor{white} InternVL3.5-8B & 1.5 & 4.4 & 0.3 & 0.0 & 7.5 & 4.5 & 0.0 & 0.0 \\
    \rowcolor{black!6} InternVL3.5-14B & 5.0 & 5.9 & 0.3 & 2.0 & 7.0 & \third{12.6} & 0.0 & 0.0 \\
    \rowcolor{white} Molmo2-4B & 2.3 & 1.0 & 0.0 & 0.0 & 7.9 & \best{21.3} & 0.6 & 0.0 \\
    \rowcolor{black!6} Molmo2-O-7B & 2.8 & 1.0 & 0.0 & 0.0 & \third{11.6} & \second{16.8} & 0.6 & 0.0 \\
    \rowcolor{white} Pixtral-12B & $-$89.4 & $-$30.6 & 0.0 & 0.0 & $-$46.9 & $-$14.5 & 0.0 & 0.0 \\
    \rowcolor{black!6} Mistral Small 3.2-24B & 2.0 & 5.9 & 0.0 & 0.0 & $-$2.0 & 5.1 & 0.6 & 0.0 \\
    \midrule
    \rowcolor{white} \multicolumn{9}{@{}l}{\textit{Human reference, Human300 sample}} \\
    \rowcolor{white} GPT-5.6 Sol (Human300) & 83.7 & 63.2 & 52.2 & 38.5 & 53.5 & 72.7 & 55.0 & 21.7 \\
    \rowcolor{black!6} \textbf{Human} (two architects) & \textbf{98.0} & \textbf{97.7} & \textbf{97.8} & \textbf{90.4} & \textbf{93.0} & \textbf{95.5} & \textbf{93.8} & \textbf{89.1} \\
    \bottomrule
  \end{tabularx}
\end{table*}

\subsection{Main Results and Human Reference}
\label{sec:human}

\Cref{tab:results} separates models that can choose from models that can localize. Sol leads seven of the eight columns, and only Sol and Terra find a substantial share of regions on their own (57.8\%/54.0\% and 40.2\%/31.3\% of Natural/BIM boxes). Gemini, Grok, and DeepSeek choose candidates well, Gemini best of all on Natural Select, yet none of them localizes more than 16.6\% of boxes. The remaining seven frontier models stay at or below 32.0 points on Judge/Select and 1.5\% on boxes, and the ten smaller open-weight models stay below 22 points and localize at most one box and one line per source; we fine-tune the strongest of them on Natural, Qwen3.5-9B, in \cref{sec:sft}. Rankings also shift with the kind of query: on BIM Select, Gemini matches Sol on cross-view matching (70.5 vs.\ 66.2 points) but clearly trails it on novel-section prediction (57.2 vs.\ 72.3).

\textbf{Same targets, different outputs.} The gap persists when the target is held fixed. On 134 Natural region pairs that share the target, reference, and base drawing, Sol, Gemini, and Grok pick the right candidate yet fail to localize it on 48, 104, and 119 pairs under the content criterion; the reverse happens only 5, 0, and 0 times (Supplement D.3).

\textbf{Section cuts.} Lines are harder still. Sol again leads, with 40.0\% on Natural and 17.5\% on BIM, well below its box accuracies. The median frontier model localizes 2.0\% and 0.0\% of lines, and no model other than Sol exceeds 5\% on BIM.

\textbf{Human reference.} Since Sol is the strongest line localizer, we compare it with the two architects on the same Human300 questions. The architects place 90.4\% of Natural and 89.1\% of BIM cuts correctly, against Sol's 38.5\% (10/26) and 21.7\% (5/23), a weighted gap of 51.9 and 67.4 percentage points (Supplement E). Coordinate output alone does not explain this gap: on a single drawing, Sol redraws 98\% of marked boxes and 94\% of marked lines (\cref{sec:controls}).

\subsection{Recognizing versus Localizing}
\label{sec:controls}

Could these failures simply reflect an inability to read drawings or to write coordinates? We test this with single-image controls for the 12 frontier models. In 495 questions, the model recognizes object categories, symbols or text, or artificial marks; in 200 more, it redraws a red box or directed line marked on a single drawing. For 190 of the redraw questions, the marked target, base drawing, GT, and scoring are shared with a cross-view question, so the two outputs can be compared target by target (Supplement F.1).

\textbf{Some failures begin in a single image.} Several models struggle with basic recognition or mark reproduction. ERNIE scores 18.8 points on Natural mark recognition, and Mistral Large 33.0/24.2 points on Natural/BIM object recognition. Kimi, Mistral Large, and Llama redraw only 8, 0, and 0 of the 100 marked boxes. For a few models, coordinate conventions explain part of this: choosing each model's coordinate unit from the controls lifts GLM's box accuracy to 14.5\%/6.1\%, although cross-view localization stays low.

\textbf{Cross-view localization adds substantial difficulty.} Pooled over the 12 models, 704 of the 2,280 model--target pairs are redrawn correctly from a visible mark, but only 187 are found across views. Qwen3.5-397B scores 88.8/80.8 points on Natural/BIM object recognition yet localizes only 1.5\%/0.6\% of boxes across views, and MiniMax redraws 61 of the 100 marked boxes but localizes one. Even on targets they redraw correctly, Sol and Terra localize only 85 of 183 (46.4\%) and 38 of 166 (22.9\%). For models that can read and redraw, much of the remaining difficulty lies in carrying the target from one view to another. The reference view matters: blanking it in 50 component-matching questions drops Sol's Select score from 78.7 to 4.0 points on Natural and from 52.0 to 4.0 on BIM, close to chance.

\textbf{Wrong instance or wrong extent.} Aligning saved predictions with annotated instances on 314 Natural door/window and 113 BIM region questions shows what the failed boxes look like. Of Sol's valid geometric errors, 24 of 149 on Natural and 34 of 52 on BIM land on another instance of the same kind, and a further 66 and 6 are centered on the target but miss its extent. Both wrong-instance matches and extent errors are thus common, and on BIM the wrong instance dominates.

\section{Error Structure}
\label{sec:analysis}

Recognition and coordinate output alone thus do not explain the localization failures. We next ask what the errors look like: where incorrect correspondences land, and which drawing constraints failed cuts violate.

\begin{figure}[t]
  \centering
  \begin{subfigure}[t]{\columnwidth}
\centering
\includegraphics[width=\linewidth]{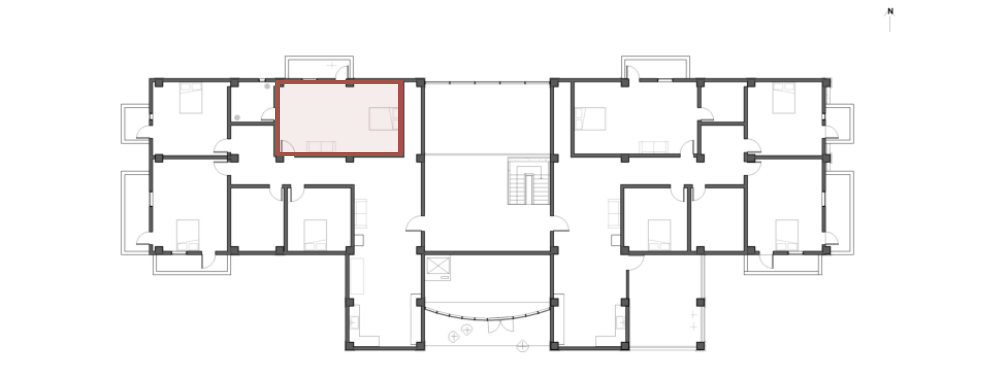}
\caption{Reference plan, floor 6}
\end{subfigure}
\par\vspace{3pt}
\begin{subfigure}[t]{\columnwidth}
\centering
\includegraphics[width=\linewidth]{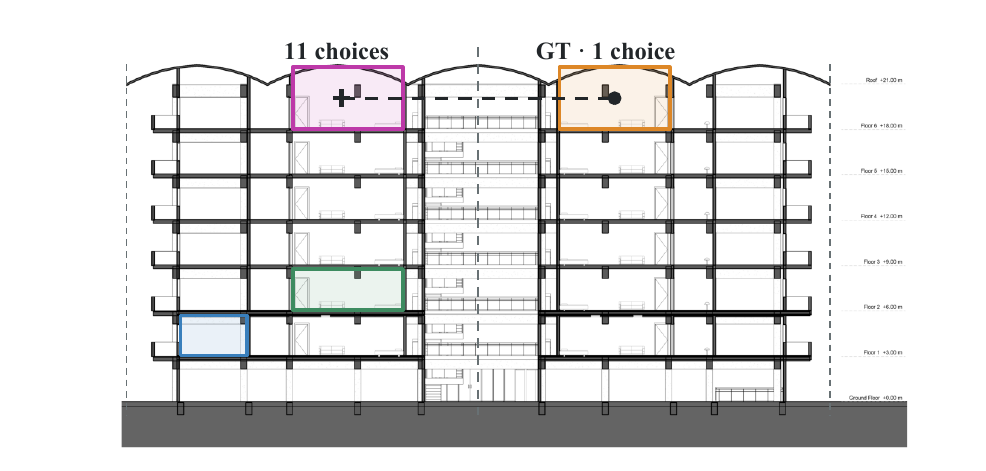}
\caption{Target section and choices}
\end{subfigure}

  \caption{Mirrored-room errors: 11 of 12 models select the room mirrored from the correct one. Circle: GT center; cross: its reflection; dashed line: target-frame midline.}
  \label{fig:wrong-choices}
\end{figure}

\begin{figure}[t]
  \centering
  \begin{subfigure}[c]{0.62\columnwidth}
    \centering
    \includegraphics[width=\linewidth]{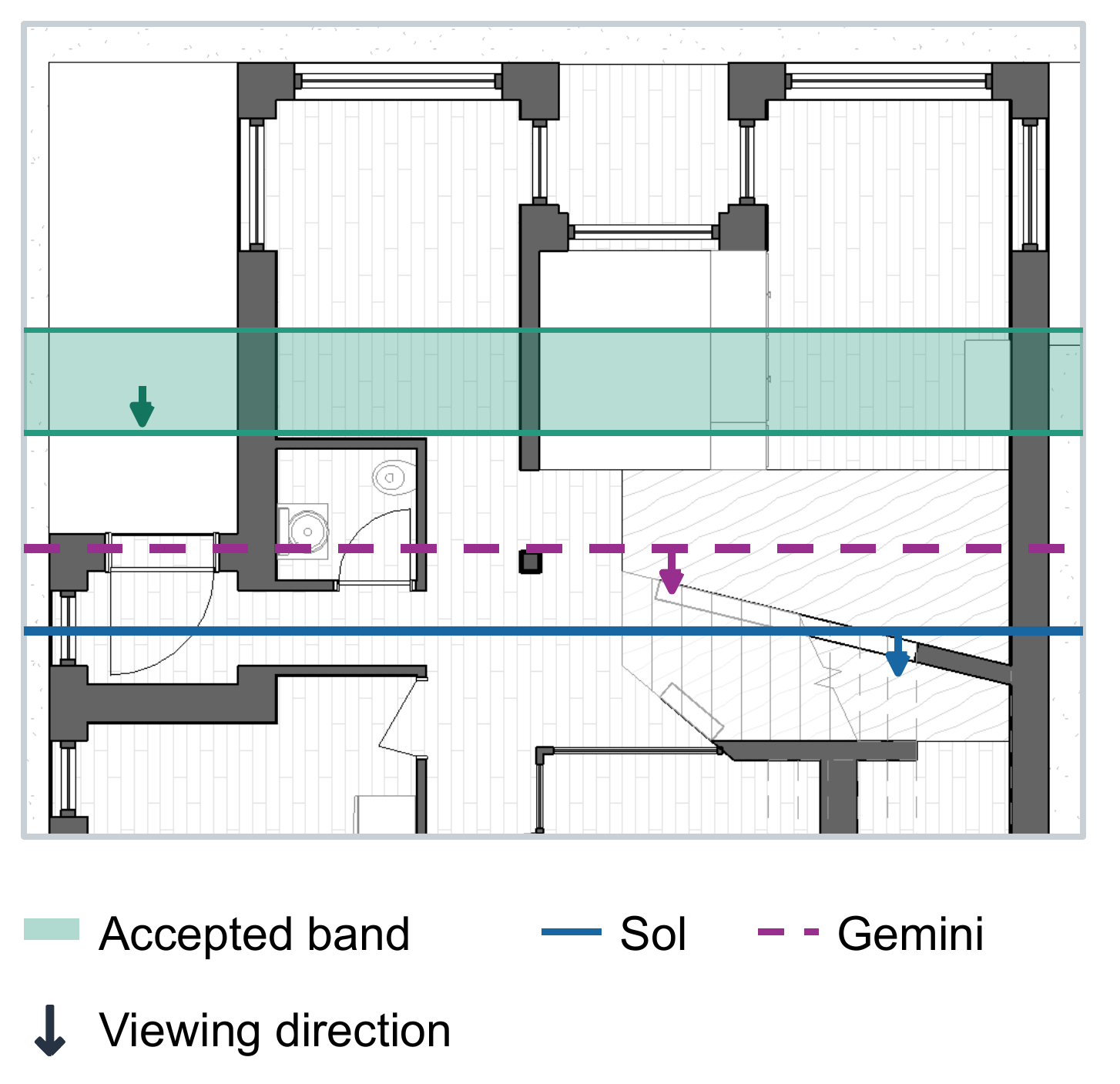}
    \caption{Inverse cut localization}
  \end{subfigure}\hfill
  \begin{subfigure}[c]{0.36\columnwidth}
    \centering
    \includegraphics[width=\linewidth]{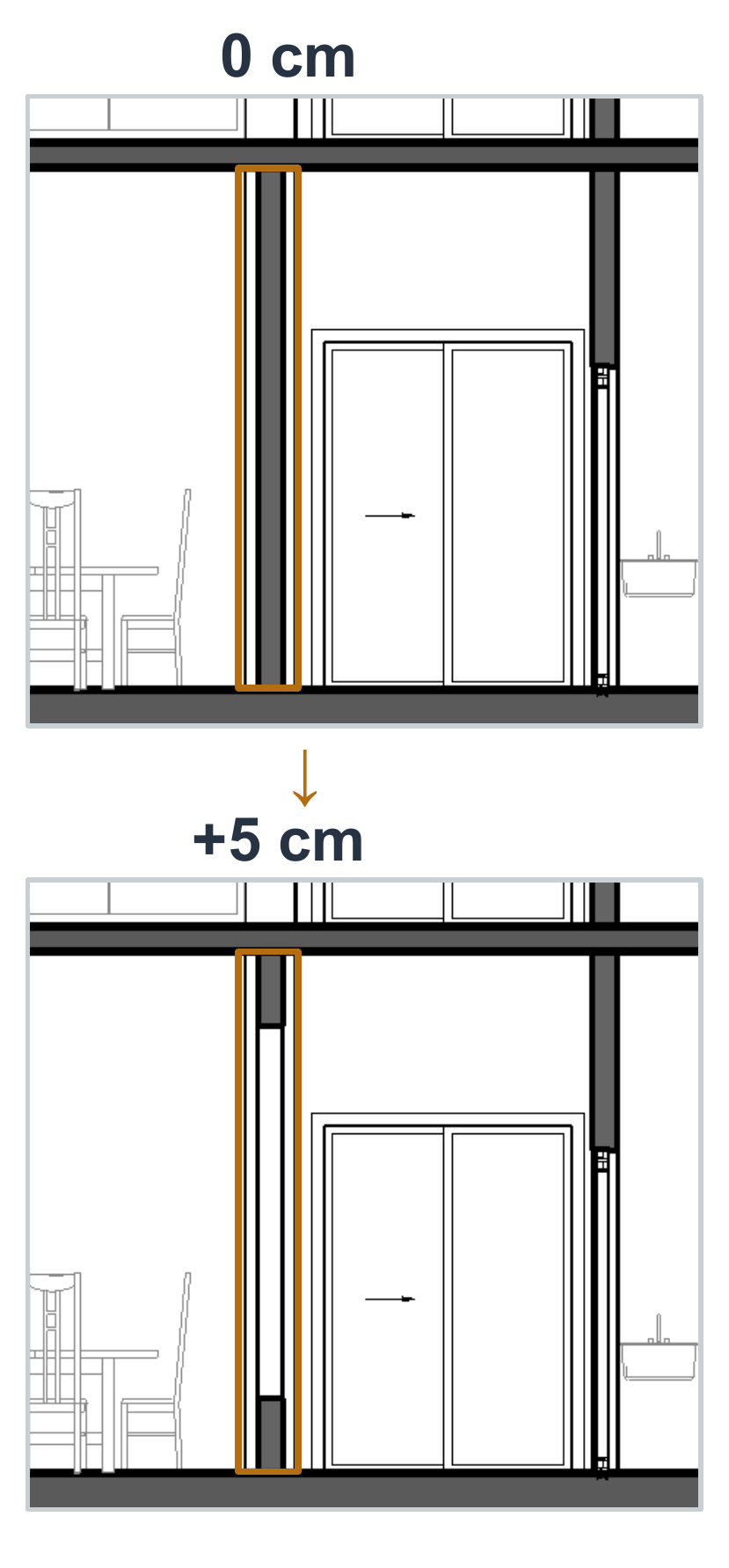}
    \caption{Fixed direction, shifted cut}
  \end{subfigure}
  \caption{Cut-position errors and local sensitivity. (a) Sol and Gemini predict the correct viewing direction but miss the accepted band. (b) In another building, a 5\,cm cut shift changes the local jamb fill.}
  \label{fig:loc-errors}
\end{figure}

\subsection{Where Incorrect Correspondences Land}
\label{sec:wrong-choices}

We analyze 900 responses from 12 models on 75 BIM matching questions across 15 buildings. Because wrong choices partly reflect preferences for option positions and colors, we measure \textbf{excess incorrect votes}, the observed incorrect votes minus those expected from these preferences, and correlate them with spatial measures of the incorrect candidates within each question. Permutation tests keep the agreement among models intact (Supplement F.4).

Wrong choices are drawn toward mirror images of the target. Candidates closer to its horizontal mirror position, or to the position obtained by reversing the shared cross-view axis, receive more excess incorrect votes ($r = 0.311$ and $0.305$, both BH $q = 0.0084$). The target's mirrored surrounding layout ($r = 0.290$) and a combined measure of shared-axis reversal and mirrored layout ($r = 0.239$) show the same tendency. The first three associations survive building-level tests after BH correction across 19 measures (all $q = 0.0425$), and the combined measure reaches $q = 0.0609$. \Cref{fig:wrong-choices}(a)--(b) shows a typical case, in which 11 of 12 models choose a similar room on the opposite side of the same storey. Near-mirrored rooms also resist a change of reference: when the reference alternates between two near-mirrored bedrooms, Sol picks the same bedroom in all 12 responses (Supplement F.2).

\subsection{Recovering the Cut from a Section}
\label{sec:loc-errors}

In the forward direction, a section is easy to describe: the cut intersects some elements and reveals others behind it. Small shifts already matter, since moving the cut by 5\,cm with fixed viewing conditions changes the jamb fill in a controlled BIM example (\cref{fig:loc-errors}(b)). The inverse direction is where models struggle. Sol reproduces 48 marked lines in Natural and 37 in BIM, but recovers only 20 and 7 of these from the reference section and the unmarked plan.

Many failed cuts get the direction right and the position wrong. Under the content criterion, 211 of 600 Natural and 142 of 480 BIM responses from the 12 models have the correct viewing direction but fail the geometric checks, and 167 and 122 of these are parallel to the band but outside it. In \cref{fig:loc-errors}(a), Sol and Gemini both face the correct way but place the cut too far toward the stair and entrance to reproduce the reference section. Viewing direction and cut position are thus separate constraints, and models often satisfy only the first (Supplement F.3).

\section{Learning from BIM Supervision}
\label{sec:sft}

Can the supervision that CrossProjection provides close part of this gap? We fine-tune Qwen3.5-9B with a two-stage curriculum that moves from closed-form answers to explicit geometry. The model first learns all 10,831 Judge/Select questions of the training buildings, and then learns region boxes and section lines while closed-form questions are replayed. Both stages use standard cross-entropy on the answer alone, with LoRA~\cite{hu2022lora} on the language side and no geometric reward or localization head, so the model must learn to localize through the same text interface it uses to judge and select.

We split the 23 BIM buildings into 15 training, four development, and four test buildings. Box and line localization is evaluated on the 2,004 free-form questions of the test buildings, with boxes scored at IoU $\geq 0.60$ as in the benchmark and lines by viewing direction and endpoint error. Judge and Select are evaluated on the BIM benchmark questions, and transfer on all 1,672 Natural benchmark questions, none of which is used for fine-tuning (Supplement G).

\begin{table}[t]
  \centering
  \small
  \setlength{\tabcolsep}{3pt}
  \caption{Curriculum fine-tuning of Qwen3.5-9B on BIM supervision. BIM Judge/Select use the benchmark questions and BIM Box + Line the four held-out buildings; Natural questions are not used for fine-tuning. Raw accuracy (\%); Box + Line pools boxes and lines. Green: gain in percentage points.}
  \label{tab:sft}
  \setlength{\aboverulesep}{0pt}\setlength{\belowrulesep}{0pt}
  \renewcommand{\arraystretch}{1.15}
  \begin{tabularx}{\columnwidth}{@{}l YYY@{}}
    \toprule
    \rowcolor{headgray} Qwen3.5-9B & Judge & Select & Box + Line \\
    \midrule
    \multicolumn{4}{@{}l}{\textit{BIM} {\footnotesize(441 / 447 / 2,004 questions)}} \\
    Original & 56.7 & 32.7 & 0.0 \\
    \rowcolor{oursrow} Fine-tuned & \textbf{62.1}\gain{+5.4} & \textbf{45.4}\gain{+12.8} & \textbf{25.0}\gain{+25.0} \\
    \midrule
    \multicolumn{4}{@{}l}{\textit{Natural, not used for fine-tuning} {\footnotesize(794 / 435 / 443 questions)}} \\
    Original & 61.0 & 35.2 & 0.5 \\
    \rowcolor{oursrow} Fine-tuned & \textbf{64.6}\gain{+3.7} & \textbf{50.1}\gain{+14.9} & \textbf{27.5}\gain{+27.1} \\
    \bottomrule
  \end{tabularx}
\end{table}

\textbf{Results.} Fine-tuning teaches the model to localize regions it could not localize before (\cref{tab:sft}). On held-out BIM buildings, it places 500 of 1,976 boxes correctly (25.3\%), against none before training, and its Select and Judge accuracies on the BIM benchmark rise by 12.8 and 5.4 points. The gains carry over to real drawing sets without any Natural supervision: Natural box accuracy rises from 0.3\% to 30.8\% (121 of 393), a level that only Sol and Terra exceed among the 22 models in \cref{tab:results}, and Select accuracy rises from 35.2\% to 50.1\%. Section lines remain the hardest output, with 2 of 28 held-out BIM cuts and 1 of 50 Natural cuts correct; they are also scarce in the supervision, with 106 training questions against 3,325 for regions.

\textbf{What transfers.} BIM views use uniform graphics, whereas the Natural sets follow the varied conventions of real practices. That box localization still transfers suggests the model learns to carry targets across views rather than BIM-specific appearance.

\section{Conclusion}
\label{sec:conclusion}

CrossProjection asks whether VLMs can turn cross-view correspondence in architectural drawings into geometric answers, from Judge and Select to region localization and section cuts scored against expert-adjudicated bands. Across 22 models, success with supplied candidates or visible marks rarely carries over to localizing the same targets in another view, and a correct viewing direction does not guarantee a cut compatible with local geometry. BIM supervision narrows the gap for regions: a fine-tuned 9B model learns to localize boxes in held-out BIM buildings and in real drawing sets not used for fine-tuning, while section lines improve only slightly. Precise section-cut localization thus remains a concrete open challenge, calling for richer section supervision than the 106 training cuts used here. We will release the benchmark, evaluation resources, and both construction pipelines.

\paragraph{Limitations.}
Our evidence comes from architectural drawings, two architects, and a single fine-tuning seed, and the spatial associations describe behavior, not causes.

\FloatBarrier

\section*{LLM Usage Statement}
We used Codex to assist with manuscript preparation, coding, analysis, illustrative graphic generation, and pre-submission feedback. Benchmark ground truth was derived from human annotations and deterministic geometry. All AI-assisted outputs were reviewed by the authors, who take full responsibility for the work.

{
    \small
    \bibliographystyle{ieeenat_fullname}
    \bibliography{main}
}

\clearpage
\twocolumn[{\centering\Large\bfseries Supplementary Material\par\vspace{1.5em}}]
\suppressfloats[t]
\appendix
\renewcommand{\topfraction}{0.9}
\renewcommand{\dbltopfraction}{0.9}
\renewcommand{\textfraction}{0.07}
\renewcommand{\floatpagefraction}{0.85}
\renewcommand{\dblfloatpagefraction}{0.85}
\setcounter{topnumber}{4}
\setcounter{dbltopnumber}{3}
\setcounter{totalnumber}{6}
\setlength{\textfloatsep}{12pt plus 2pt minus 2pt}
\setlength{\dbltextfloatsep}{12pt plus 2pt minus 2pt}
\setcounter{table}{0}
\setcounter{figure}{0}
\setcounter{equation}{0}
\renewcommand{\thetable}{S\arabic{table}}
\renewcommand{\thefigure}{S\arabic{figure}}
\renewcommand{\theequation}{S\arabic{equation}}
\renewcommand{\theHtable}{S\arabic{table}}
\renewcommand{\theHfigure}{S\arabic{figure}}
\renewcommand{\theHequation}{S\arabic{equation}}

\noindent This appendix gives the full methods and results behind the main paper. Pointers such as ``Sec.~4.2 of the main paper'' refer to the main text; all other section, figure, and table numbers refer to this appendix.

\section{Task Definitions and Complete Examples}
\label{sup:a}

Sec.~3.1 of the main paper defines the tasks only briefly. Here we explain what each task asks of a solver and how information in one view constrains the answer in another.

\subsection{2D Representations of 3D Space and the Abilities Tested}
\label{sup:a1}

Plans, elevations, and sections each show only part of a building's 3D structure, and they do so through projection, sectioning, and drafting symbols. The viewing direction sets the axes and decides what counts as front and back, the cut position decides which elements are sliced, and drafting conventions decide which details are drawn at all. One entity can therefore look quite different from drawing to drawing, while two entities that look alike may be unrelated.

To decide which abilities the tasks should probe, we follow the five spatial abilities of Xu \etal~\cite[Tab.~1, Secs.~3.1--3.2]{xu2025basic}. Spatial Perception lets a reader recognize a drawing's orientation and geometry, and Spatial Relation lets them link entities and their local relations. Spatial Orientation is needed whenever the viewing side or reference frame changes, Mental Rotation whenever the shape or correspondence of a rotated object must be inferred, and Spatial Visualization whenever projection, occlusion, or a sectioned structure must be imagined.

\Cref{tab:S1} turns these abilities into concrete checks by pairing recurring situations in drawings with the spatial operation each one demands.

\begin{table}[t]
  \centering
  \footnotesize
  \setlength{\tabcolsep}{3pt}
  \caption{Drawing conditions and spatial operations.}
  \label{tab:S1}
  \begin{tabularx}{\columnwidth}{@{}LL@{}}
    \toprule
    Condition in the drawings & Spatial operation to be checked \\
    \midrule
    The same component appears with different outlines and symbols in plan, elevation, or section & Establish entity correspondence rather than merely recognizing the same category \\
    The viewing side, projection axis, or cutting condition changes & Transform positional and directional relations into the applicable reference frame, and decide which relations are comparable across drawings \\
    Similar openings, rooms, or components occur repeatedly & Combine floor, adjacency, arrangement, and reliable occlusion relations to discriminate candidate correspondences \\
    \bottomrule
  \end{tabularx}
\end{table}

\textbf{From spatial abilities to the three task categories.} Each category leans on a different ability. Cross-view Matching turns spatial relations into the question of whether two views show the same entity or space. Projection Registration adds spatial orientation, because the solver must locate one view's viewing direction and reference frame within another. Novel-section Prediction rests mainly on spatial visualization: the solver must imagine what a specified cutting plane would reveal. Spatial perception underlies all three, and mental rotation enters when the particular views and object transformations call for it.

\begin{figure*}[t]
  \centering
  \includegraphics[width=\textwidth]{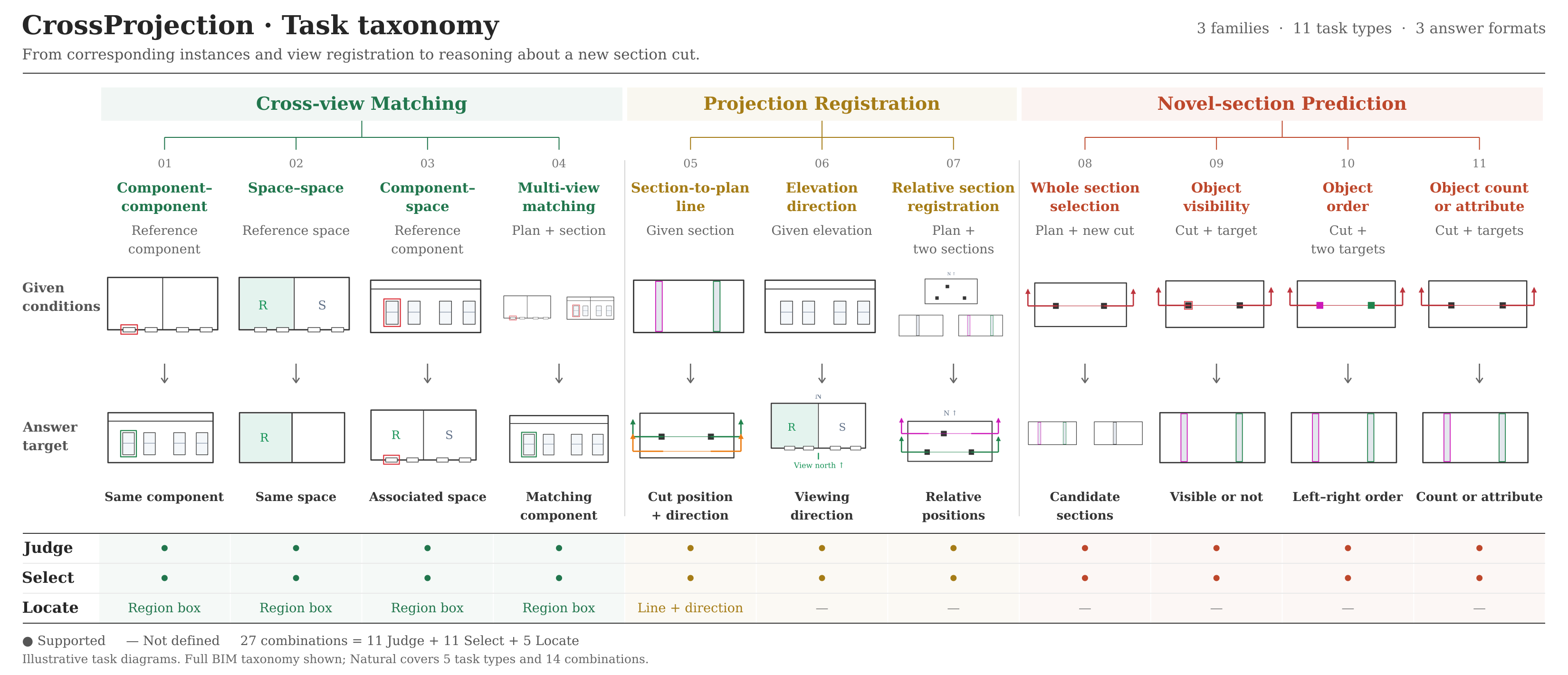}
  \caption{Schematic examples of the 11 task subtypes. Each column reads from the given conditions to the answer target; the bottom rows show the supported answer formats. The full BIM taxonomy is shown (27 subtype--format combinations); Natural covers five subtypes and 14 combinations.}
  \label{fig:taxonomy}
\end{figure*}

\textbf{From three categories to 11 subtypes.} We then split each category by what is being related. Matching has four subtypes, defined by the objects and view combinations involved; registration has three, defined by what is registered; and section prediction has four, defined by the content to be predicted. \Cref{tab:A1-1} lists the operation behind each subtype and how it draws on each ability (SP, SR, SO, MR, and SV abbreviate the five abilities above). ``Core'' marks central involvement, ``Basis'' the perceptual foundation, and ``Varies'' involvement that depends on the view relation and question cues. \Cref{fig:taxonomy} shows what each subtype gives and what it asks for. Section-to-Plan Line (05) and Whole Section Selection (08), for example, are inverses: the first asks where and in which direction a given section was cut, while the second fixes the cut and asks which candidate section it produces.

\begin{table*}[t]
  \centering
  \footnotesize
  \setlength{\tabcolsep}{3pt}
  \caption{Spatial abilities and spatial operations in the 11 task subtypes. Core: central involvement; Basis: perceptual foundation; Varies: depends on the view relation and question cues.}
  \label{tab:A1-1}
  \begin{tabularx}{\textwidth}{@{}p{0.15\textwidth}p{0.25\textwidth}ccccc L@{}}
    \toprule
    Task category & Subtype & SP & SR & SO & MR & SV & Spatial operation \\
    \midrule
    Cross-view Matching & Component--Component & Basis & Core & Varies & Varies & Varies & Identify the same entity from its position and neighboring structure \\
    Cross-view Matching & Space--Space & Basis & Core & Varies & Varies & Varies & Match spaces using containment, adjacency, floor, and layout \\
    Cross-view Matching & Component--Space & Basis & Core & Varies & Varies & Varies & Determine the relation between an entity and the space it belongs to or adjoins \\
    Cross-view Matching & Multi-view & Basis & Core & Varies & Varies & Varies & Maintain the identity of the same target across multiple views \\
    \midrule
    Projection Registration & Section-to-Plan Line & Basis & Core & Core & Varies & Core & Infer the plan cut position and viewing direction from the section structure \\
    Projection Registration & Elevation Direction & Basis & Varies & Core & Varies & Varies & Map the elevation viewing side to the plan reference frame \\
    Projection Registration & Relative Section Registration & Basis & Core & Core & Varies & Varies & Compare cut positions and relative directions \\
    \midrule
    Novel-section Prediction & Whole Section Selection & Basis & Varies & Varies & Varies & Core & Infer the overall section depiction under a specified cutting condition \\
    Novel-section Prediction & Object Visibility & Basis & Varies & Varies & Varies & Core & Decide whether an object appears under a given cutting plane and viewing side \\
    Novel-section Prediction & Object Order & Basis & Core & Core & Varies & Varies & Convert an object arrangement into its drawing order under the viewing direction \\
    Novel-section Prediction & Object Count or Attribute & Basis & Varies & Varies & Varies & Varies & Determine a count or attribute from sectioning, visibility, and drawing information \\
    \bottomrule
  \end{tabularx}
\end{table*}

Because every question draws on several abilities at once, we report scores by task subtype and answer format rather than by ability.

\subsection{Task Inputs and Reference Dependence}
\label{sup:a2}

Each question supplies drawings, textual conditions, and, where relevant, candidates; marks drawn on the images are part of the input. BIM identities, 3D geometry, and the GT are used only to build and verify questions and are never shown to the solver. Consider ``locate in the plan the door marked in the elevation.'' The reference drawing names the object, the target drawing supplies the candidate entities and their coordinates, and only cross-view relations connect the two. Moving the reference mark to another door while leaving the target side untouched changes the correct answer, so the question cannot be solved from the target alone. \Cref{sup:f1-reference-ablation,sup:f2} test this dependence directly, by blanking the reference and by switching it.

\subsection{Cross-view Reasoning}
\label{sup:a3}

Answering such questions takes both projection geometry and drafting knowledge. Geometrically, an orthographic elevation viewed from side $v$ maps a plan position $p$ to the horizontal coordinate $u_v(p)=s\,t_v^{\mathsf T}p+b$ ($s>0$), where $t_v$ points to the right of the elevation; changing the viewing side therefore reverses left--right order and front--back relations. Drafting knowledge is needed to read occlusion, door leaves, glazing, and upper structures correctly. In practice, a solver proposes a viewing direction or correspondence, predicts what the drawing should then show, and revises the hypothesis when the prediction conflicts with what is drawn.

\subsection{Answer Identifiability}
\label{sup:a4}

The two kinds of answers differ in how tightly the drawings pin them down. For entity questions, the public drawings determine a unique answer. For section localization, several cut positions can produce the same drawing, so we accept a positional band rather than a single line (\cref{sup:b1-line-band,sup:c2-line-band}).

\subsection{Worked Cross-view Interpretation of TN House}
\label{sup:a5}

\begin{figure*}[t]
  \centering
  \includegraphics[width=\textwidth]{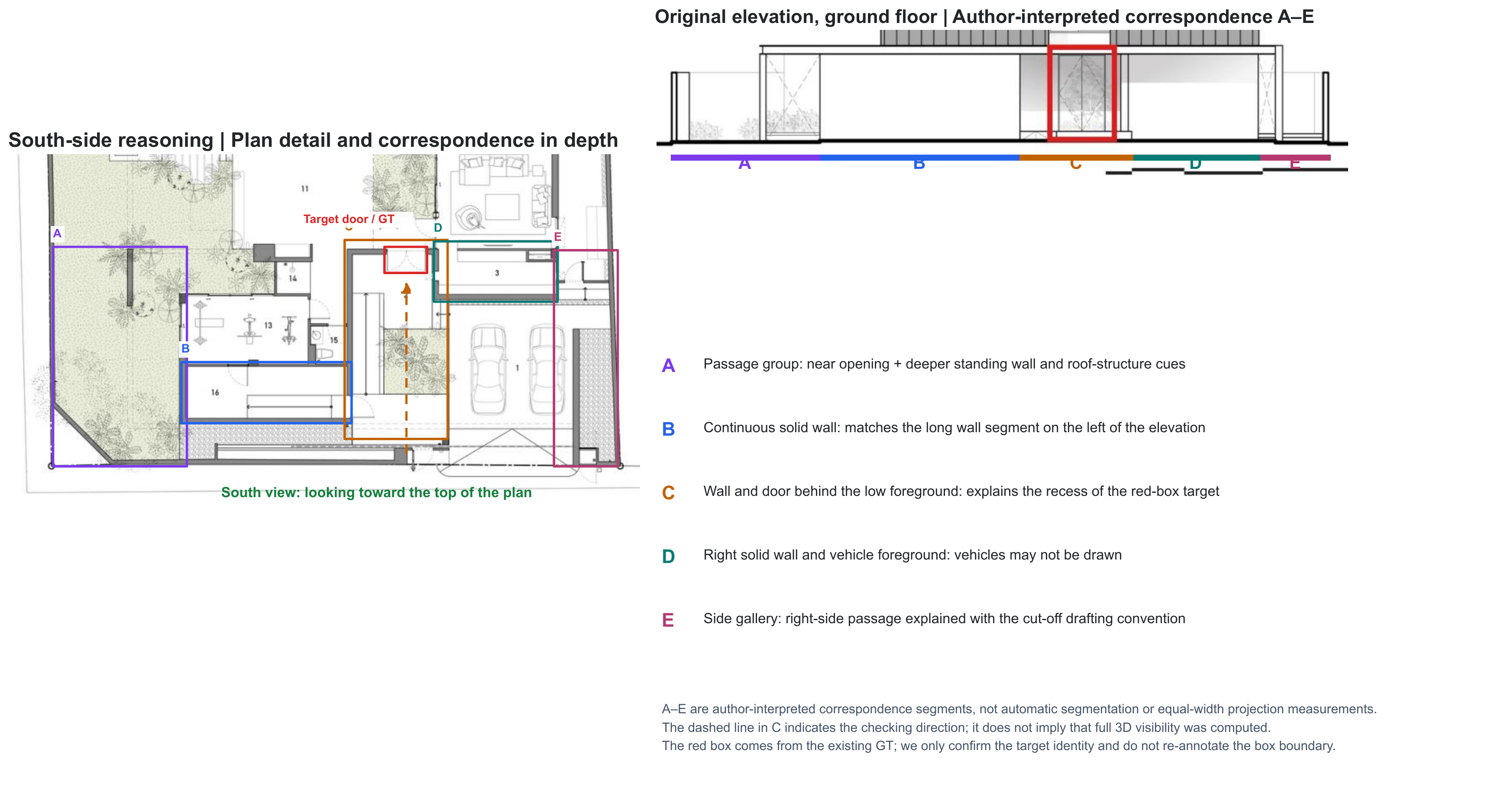}
  \caption{Cross-view correspondence for TN House. A--E are the correspondence regions marked by the authors. The red box in the elevation designates the queried object, and the red box in the plan shows the annotated target. ``South side'' means looking upward from the bottom of the plan.}
  \label{fig:S1}
\end{figure*}

\Cref{fig:S1} walks through this hypothesize-and-check process on a component localization question from TN House. The red box in the reference elevation designates the object, which must be found in the ground-floor plan. Comparing the drawings, the authors first read the perimeter wall as omitted or partially cut away and the main building as unsectioned, which confines the correspondence to the ground floor. Starting from the plan, they then predict the sequence of passages, solid walls, and openings that each viewing side should show in the elevation.

Three of the four sides can be ruled out. A west view would show a long glazed facade, yet the glazed segment in the elevation is shorter, and no occlusion accounts for the difference. A north view would not reproduce the widths of the two side passages or the double-passage combination. An east view would show a continuous solid wall at the center, which contradicts the openings drawn in the elevation.

Only the south view fits: its predicted sequence of double passage, solid wall, recessed central opening, solid wall, and side passage matches segments A--E in the drawing. Reading in depth along segment~C, the authors then identify the opening behind the low foreground as the door between the entrance hall and the terrace. The two constraints work together here: the viewing direction narrows the entity correspondence, and the depth relation in turn pins down the position in the plan.

\subsection{Cross-domain Comparison and Positioning among Neighboring Measurements}
\label{sup:a6}
\label{sup:a6-cross-domain}
\label{sup:fig-s14-cross-domain}

Tab.~1 of the main paper compares benchmarks by task coverage and by how their GT is constructed. Coverage follows each benchmark's evaluation tasks; for SPAR, we use SPAR-Bench rather than the SPAR-7M training data. The columns are defined as follows. ``Multi-image/3D input'' includes multiple images, multiple views on one sheet, and 3D volumes. ``Projection/section reasoning'' includes orthographic projection relations or cross-section inference. ``Judge/Select'' refers to answers from fixed categories or options. ``Free-form localization'' refers to directly outputting points, boxes, or lines scored by geometric position. \emph{Human} means that people directly provide the original reference answers or target annotations; verifying proposals generated by a model does not count. \emph{Geometric} means deterministic derivation from geometric relations. \emph{Dep.} means that, in at least part of the construction pipeline, a generative model takes part in forming reference answers or target annotations, including proposals that people later verify. A generative model used only to phrase questions or to score responses does not count toward Dep. Human annotation, geometric derivation, and generative-model participation can coexist within one benchmark. A dash means that the paper does not clearly document a construction method meeting the column's definition; it does not imply the absence of human involvement. The four drawing and cross-view benchmarks below mix several of these sources, so we explain how each was marked.

\textbf{AECV-Bench.} The counting task uses human annotations. For the drawing question-answering set, GPT-5 first proposes candidate question--answer pairs, which human reviewers audit, filter, and supplement with manually written pairs. Tab.~1 therefore marks both Human and Dep.~\cite[Sec.~3.2.1, p.~9]{kondratenko2026aecv}.

\textbf{MechVQA.} Two construction paths form reference answers through answers from multiple models and semantic majority voting; another path uses expert-verified metadata and expert-constructed questions with known answers. The benchmark therefore combines generative-model participation with directly human-provided GT, and Tab.~1 marks both Human and Dep.~\cite[Sec.~3.3, pp.~4--5]{kou2026mechvqa}.

\textbf{LED-BENCH.} A language model proposes candidate answers, and annotators with engineering backgrounds check the retained questions and reference answers against the original drawings, revising or discarding unsuitable candidates. We mark Dep.\ because the model takes part in forming the reference answers; the subsequent human verification does not meet our definition of Human~\cite[Sec.~3.2, p.~4; App.~F, p.~14]{li2026ledbench}.

\textbf{CrossPoint-Bench.} The construction pipeline combines Gemini-assisted annotation of actionable regions, human verification of boxes, and geometric computation of cross-view corresponding point coordinates; the evaluation set follows the same pipeline and is verified by people. Dep.\ reflects generative-model participation in target annotation, and Geometric reflects the derivation of coordinates; the human verification alone does not meet our definition of Human~\cite[App.~A.2--A.3, pp.~14--15]{wang2025crosspoint}.

\section{Data Construction, Review, and Release}
\label{sup:b}

This section explains where the data come from, how questions are built and checked, and what will be released.

\subsection{Sources and Full Support Matrix}
\label{sup:b1}

The two sources serve complementary purposes. Natural keeps architectural drawings in their native form, with the full variety of real practice in drafting style, text annotation, layout, and convention; its drawings come from ArchDaily, FreeFarmhouse, and YourHome. BIM instead generates standardized plans, elevations, and sections in Revit from building models that retain entity identity, so cross-view component correspondence, view geometry, and some spatial relations can be derived from the model itself. GNI and Library Revit (LR) are the two BIM source series, and Fig.~3 of the main paper shows both construction pipelines.

In total, the question bank contains 33,788 questions from 46 buildings/drawing sets (Natural 3,380/23, BIM 30,408/23), from which we fix an evaluation subset of 2,763 questions from 38 buildings/drawing sets (Natural 1,672/23, BIM 1,091/15). Within BIM, GNI contributes 15,456 questions from 11 buildings and LR 14,952 questions from 12 buildings to the full bank, and 533 questions / 7 buildings and 558 questions / 8 buildings to the evaluation subset.

\paragraph{Task design and geometric annotation.}
Not every subtype supports every answer format (\cref{fig:taxonomy}). Cross-view Matching supports Judge, Select, and region Locate, and Section-to-Plan Line supports Judge, Select, and line Locate; the remaining registration and section-prediction subtypes support only Judge and Select. This yields 27 valid ``subtype $\times$ answer format'' combinations in BIM. Natural, whose annotations are manual, covers five subtypes (component correspondence, space correspondence, multi-view, section-line registration, and elevation direction) and 14 combinations. Novel-section Prediction asks about the content of a new section through selection or relational judgment; models never have to generate a section image.

The Natural annotations were produced by two architects, each with ten years of professional experience, who worked together: they examined each drawing set, discussed cross-view correspondences and section relations, and agreed on the regions, anchors, and section lines to mark. For BIM, the same information comes from the model: entities, views, and each entity's occurrence in each view, linked to drawing regions, room outlines, and available visibility data. For structural columns, visibility is decided from Element-ID masks rather than from bounding boxes alone.

Question generation then turns these annotations into questions. For each task, it assembles the required views, query conditions, and answer format, draws the reference marks and candidates, and stores each answer together with its geometric basis. Locate is scored against this GT geometry, not against the reference mark, which is enlarged for display.

\paragraph{Accepted positional band constrained by section relations.}
\label{sup:b1-line-band}
Section-line questions need separate treatment because, as noted in \cref{sup:a4}, more than one cut can be correct. Two experts therefore each adjudicated all 90 section-line localization questions (50 Natural, 40 BIM) from the questions and drawings alone, without seeing the preset answers, each other's annotations, or any model or human responses. Both followed the three steps below. The main evaluation uses the first expert's bands, and the second set serves to compare boundaries and scoring outcomes (\cref{sup:d5-line-band-reannotation}).

\begin{enumerate}
  \item \textbf{Establish cross-view correspondence.} Align the corresponding regions of the reference section and the target plan according to the specified floor, spatial layout, and viewing direction.

  \item \textbf{Determine the accepted range.} With the target floor and viewing direction fixed, treat a candidate cut position $t$ on the plan as a hypothesis, infer the local geometry that a cut there would produce, and compare it with the reference section, in the same hypothesize-and-check manner as \crefrange{sup:a3}{sup:a5}. Under a single drafting interpretation, let $S_i$ be the set of cut positions that reproduce each verifiable local feature $i$ of the reference section. The accepted positions are those that satisfy every constraint:
  \begin{equation}
    S=\bigcap_i S_i.
  \end{equation}
  To place the boundaries, the experts examined which relation would change just beyond each one and whether any disconnected compatible positions existed. In all 90 questions, the compatible positions formed a single contiguous band.

  \item \textbf{Annotate the band.} The experts mark, in the coordinates of the target image, the two parallel boundaries, the evaluation span, and the viewing direction. The model outputs a single line and a direction; scoring is described in \cref{sup:c2-line-band}.
\end{enumerate}

Treating the answer as a range rather than a single line reflects what the drawings actually support. A similar idea appears in object pose evaluation, where BOP-Distrib~\cite{meden2026bopdistrib} derives pose-distribution GT from per-image visibility; we apply range annotation to registering cutting planes between architectural plans and sections. The two adjudication cases below and the controlled sweep in \cref{sup:f9} show how local geometry bounds the cut position.

\paragraph{Manual adjudication examples of accepted bands.}
\label{sup:b1-line-band-cases}
The first case involves a stair and a local object (Natural); the second, an asymmetric section at a window edge (BIM).

\textbf{Natural: stair cross-section and a local object (N03, ad\_136).} The reference section clearly shows the stair as cut, so the cut line must first cross the corresponding stair flight in the plan. On its left, the section also shows the cut outline of a local object; matching it to the same object in the plan narrows the line to positions that cross both the stair flight and that object. Together, these two constraints set the upper and lower boundaries of the band. The toilet and bed, drawn uncut in the section, help confirm the viewing direction and on which side of the cutting plane the room objects lie.

\begin{figure*}[t]
  \centering
  \includegraphics[width=\textwidth]{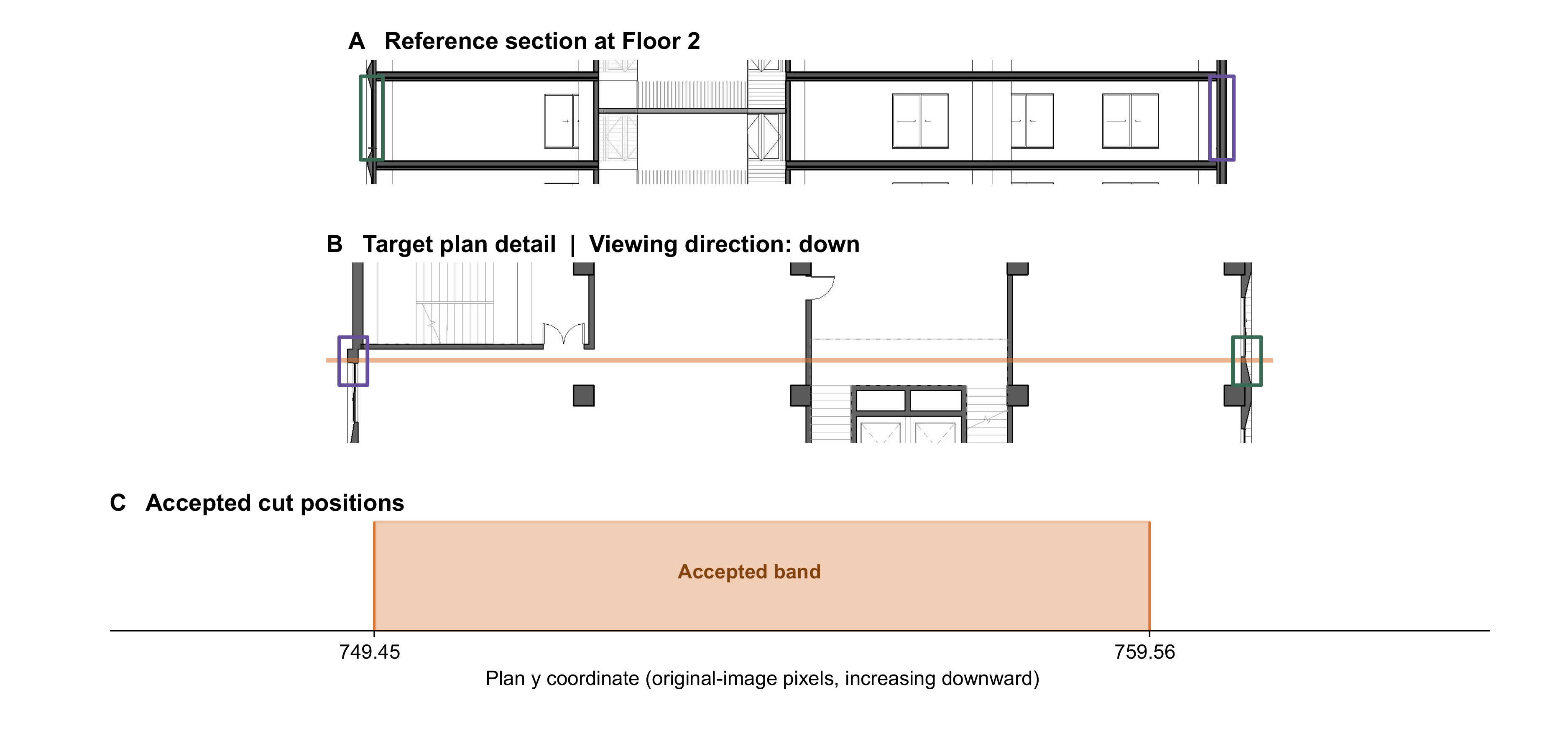}
  \caption{Accepted positional band for the B27 example. \textbf{A}: reference section. \textbf{B}: target plan. Boxes of the same color mark corresponding local evidence for a downward viewing direction. \textbf{C}: magnified interval of admissible cut positions in the plan.}
  \label{fig:S18}
  \label{sup:fig-b27-accepted-band}
\end{figure*}

\textbf{BIM: asymmetric cross-section at a window edge (B27, LR009).} Here the reference section is thin at its left end and a thicker solid at its right. Because the window-edge construction thickens gradually along the plan, the section width depends on where the cut falls. Once both ends are mapped to the plan using the viewing direction, the thin end confines the cut to a narrow region at the window edge, while the thick end requires the same line to pass through the solid part at the other end. Only the narrow band in \cref{fig:S18} satisfies both.

\textbf{Band scoring example.} On the original B27 plan, this band spans $749.45 \leq y \leq 759.56$ pixels (rounded) with a downward viewing direction, and lines on either boundary count as inside.

\paragraph{Prompt templates and candidate encoding.}
\label{sup:b1-prompt-encoding}
Each subtype and answer format has a fixed English template. Its public fields are the image indices, reference marks, candidates, and any floor conditions needed, and the required output follows the contract in \cref{sup:c2-output-contract}. \Cref{tab:B1-1} gives representative task sentences for instance correspondence and cutting-plane registration.

\begin{table}[t]
  \centering
  \footnotesize
  \setlength{\tabcolsep}{3pt}
  \caption{Representative task sentences by answer format.}
  \label{tab:B1-1}
  \begin{tabularx}{\columnwidth}{@{}>{\raggedright\arraybackslash}p{0.15\columnwidth}L@{}}
    \toprule
    Format & Representative task sentence \\
    \midrule
    Judge & Do the \{m1\} in Image 1 and the \{m2\} in Image 2 mark the same building component? \\
    Select & Which of the \{candidates\} in Image 2 marks the same building component as the \{m1\} in Image 1? \\
    Locate & In Image 2, draw a bounding box around the building component marked by the \{m1\} in Image 1. \\
    Section-line Locate & In Image \{plan\}, draw the section-cut line that produces the section shown in Image \{section\}. Also specify the viewing direction from the line toward the viewed side, using up, down, left, or right in Image \{plan\}'s page coordinates. \\
    \bottomrule
  \end{tabularx}
\end{table}

\textbf{Candidate entities, colors, and answer indices.} A candidate's color encodes its identity, and neither color nor position should hint at the answer. Natural matching questions therefore shuffle candidates under a fixed seed and rotate the label position of the correct target, and section-line selection questions likewise rotate the correct position and shuffle the rest. BIM permutes candidate entities or section lines over color labels per question and computes the correct index from the permuted position; whole-section selection questions permute both the candidate sections and the image order. Options with fixed semantics, such as Yes/No, directions, and counts, keep their natural order, and the correct one is set by the relation or geometry of the question. \Cref{sup:d1} reports majority-label baselines, and \cref{sup:f2} tests whether recoloring or reordering the candidates changes model answers.

\textbf{Evaluation subset selection.} The evaluation subset is drawn with per-source quotas and a fixed seed, balancing answer categories within each ``subtype $\times$ answer format'' cell where possible while keeping building coverage. Natural questions are sampled from all 23 drawing sets. BIM questions are sampled from 15 of the 23 buildings, 7 from GNI and 8 from LR, which together cover all 11 subtypes and 27 valid ``subtype $\times$ answer format'' combinations. This gives a 2,765-question pool (Natural 1,674, BIM 1,091), which also serves as the Human300 sampling frame (\cref{sup:e1}). Finally, two Natural line Select questions are removed because their candidates are incompatible with the accepted cut positions; their Locate counterparts are kept. Every model answers the resulting 2,763 questions (Natural 1,672, BIM 1,091).

\paragraph{BIM drawing and cross-projection supervision workflow.}
\label{sup:b1-bim-workflow}
The BIM workflow turns 3D models into drawings, entity correspondences, geometric annotations, and cross-projection questions. GNI and LR share the same pipeline, with views, floors, and cut conditions set per building.

\textbf{Inputs.} The workflow uses Revit 2026 and AEC Model Bridge 1.0.2 and reads RVT models, or IFC models imported into Revit.

\textbf{Generation procedure.}
\begin{enumerate}
  \item \textbf{Generate and check drawings.} Plans, elevations, and sections are generated or refined for each building with unified display and canvas settings, after which the drawings to adopt are confirmed manually.
  \item \textbf{Extract correspondence data.} The extraction step retains entity identity, the view projection coordinate system, the model-to-pixel mapping, visible vectors, Element-ID pixel ownership, and available room regions; these data support cross-view entity association and the construction of geometric answers. Component bounding boxes, visible pixels, and room boundaries are kept distinct.
  \item \textbf{Form and review annotations.} The confirmed drawings are bound to the extracted data, and the views, entity correspondences, target extents, and recognizability on the public drawings are verified.
  \item \textbf{Generate questions and supervision.} Questions are generated by building, subtype, and answer format, and their candidates, drawing marks, and geometric answers are reviewed.
\end{enumerate}

\subsection{Automated Checks and Expert Adjudication}
\label{sup:b2}

Every question passes two layers of checking. Automated checks verify question parsing, canvas coordinates, and the binding between images and text. Professional review then asks whether the reference is recognizable, the target visible, and the drawing information sufficient, and whether any valid alternative answer exists.

Two architectural experts, neither of whom answered Human300, jointly reviewed all 300 Human300 questions in depth, adjudicated their answers, and then screened the entire bank for analogous issues using the generation rules. A further 100 questions outside Human300, drawn with a fixed seed to span all 38 buildings and 41 valid ``subtype $\times$ answer format'' combinations, were spot-checked and showed no substantive issue. The band adjudication of all 90 line questions is described in \cref{sup:b1-line-band}.

\label{sup:b2-adjudicated-answers}
\textbf{Adjudicated answers.} Models and experts are always compared against the same adjudicated answers, which follow a few fixed conventions. For split-level and multi-floor questions, the stem states the necessary elevation. A complete space is bounded by walls and explicit partitions; wall segments with door openings still count as boundary, and connected areas without explicit partitions are treated as one space.

\subsection{Resource Composition and Release Scope}
\label{sup:b3}
\label{sup:b3-delivery-plan}

Release terms follow each source's license. Natural drawings from FreeFarmhouse (1,690 questions) and YourHome (565) are redistributed with attribution under their CC licenses, whereas ArchDaily drawings (1,125) are referenced by URL rather than mirrored. BIM questions from GNI (15,456) and LR (14,952) are released as 2D drawings with annotations and source notes; the original model files are not included. We will publicly release the questions, annotations, adjudicated GT, scorer, model responses, annotation tools, and question-generation code.

\section{Model and Scoring Protocol}
\label{sup:c}

This section specifies how models are queried, what output they must produce, and how that output is scored.

\subsection{Models and Inference Configuration}
\label{sup:c1}

All 12 frontier models answer the same 2,763 questions, for 33,156 model--question units. Each question is sent as an independent request containing the drawings, the question stem, and the answer format; \cref{tab:S5} lists platforms, dates, and explicit settings. To see how smaller models fare, we also run ten open-weight models from 4B to 24B parameters on the same questions, with the same output contract and scoring. They run locally from public checkpoints with their native processors and instruction templates (\cref{tab:S5b}), each on a single NVIDIA A100 PCIe GPU (40\,GB; 80\,GB for Mistral Small 3.2), with BF16 weights, no quantization, PyTorch 2.13, and Transformers 5.16.1 (4.57.1 for Molmo2). Thinking modes are off. Images are passed at original resolution in the declared order and resized or tiled by each native processor. Qwen3.5-9B uses the same decoding settings as the fine-tuning evaluation (\cref{sup:g1}), and all ten results appear in \cref{tab:open-weight}.

\begin{table}[t]
  \centering
  \footnotesize
  \setlength{\tabcolsep}{2pt}
  \caption{Decoding and image settings of the ten locally run open-weight models. Sampling parameters follow each checkpoint's released generation configuration. Image budgets apply per source image after native resizing or tiling.}
  \label{tab:S5b}
  \begin{tabularx}{\columnwidth}{@{}l>{\raggedright\arraybackslash}Xr>{\raggedright\arraybackslash}X@{}}
    \toprule
    Model & Decoding & Max new tokens & Image budget \\
    \midrule
    Qwen3.5-9B & Greedy & 256 & $\leq$2,007,040 px \\
    Qwen3-VL-4B & $T{=}0.7$, top-$p$ 0.8, top-$k$ 20 & 32,768 & $\leq$2,007,040 px \\
    Qwen3-VL-8B & $T{=}0.7$, top-$p$ 0.8, top-$k$ 20 & 32,768 & $\leq$2,007,040 px \\
    Qwen2.5-VL-7B & $T{=}10^{-6}$ & 32,768 & $\leq$2,007,040 px \\
    InternVL3.5-8B & Greedy & 16,384 & $\leq$9 tiles + thumbnail, 448 px \\
    InternVL3.5-14B & Greedy & 16,384 & $\leq$9 tiles + thumbnail, 448 px \\
    Molmo2-4B & Greedy & 16,384 & $\leq$8 crops + overview, 378 px \\
    Molmo2-O-7B & Greedy & 32,768 & $\leq$8 crops + overview, 378 px \\
    Pixtral-12B & Greedy & 32,768 & Longest side 1,024 px \\
    Mistral Small 3.2 & $T{=}0.15$ & 32,768 & Longest side 1,400 px \\
    \bottomrule
  \end{tabularx}
\end{table}

\begin{table*}[t]
  \centering
  \footnotesize
  \setlength{\tabcolsep}{3pt}
  \caption{Models, platforms, evaluation dates, and explicit settings. Output caps are in tokens; parentheses give request counts when caps vary. Unlisted reasoning and sampling parameters use service defaults. Effort levels are service-specific.}
  \label{tab:S5}
  \begin{tabularx}{\textwidth}{@{}llcL@{}}
    \toprule
    Model & Provider/platform & Evaluation dates (Beijing time) & Explicit settings and output caps \\
    \midrule
    GPT-5.6 Terra & OpenAI/Codex Responses & 2026-09-09--2026-10-22 & effort=high; summary=auto; output cap not explicitly set \\
    GPT-5.6 Sol & OpenAI/Codex Responses & 2026-09-09--2026-10-22 & effort=high; summary=auto; output cap not explicitly set \\
    Gemini 3.1 Pro (preview) & Google Vertex AI & 2026-09-09--2026-10-22 & thinkingLevel=HIGH; output cap: 32,768 (2,762); 8,192 (1) \\
    Grok-4.6 & Google Vertex AI & 2026-09-09--2026-10-22 & reasoning\_effort=high; output cap: 32,768 \\
    DeepSeek-V4.1-Flash & Tencent TokenHub & 2026-09-09--2026-10-22 & Output cap: 32,768 (2,761); 8,192 (2) \\
    Qwen3.5-397B-A17B & Paratera & 2026-09-09--2026-10-24 & Output cap: 32,768 (2,762); 8,192 (1) \\
    GLM-4.6V & Paratera & 2026-09-09--2026-10-21 & Output cap: 16,384 (2,762); 8,192 (1) \\
    Kimi-K2.5 & Paratera & 2026-09-09--2026-10-24 & Output cap: 32,768 (2,762); 8,192 (1) \\
    ERNIE-4.5-Turbo-VL & Paratera & 2026-09-09--2026-10-21 & Output cap: 8,192 (2,762); 2,048 (1) \\
    Mistral Large 3 & Azure Foundry & 2026-09-09--2026-10-21 & temperature=0; output cap: 8,192 (2,762); 2,048 (1) \\
    Llama 4 Maverick & Azure Foundry & 2026-09-09--2026-11-01 & temperature=0; output cap: 8,192 (2,762); 2,048 (1) \\
    MiniMax-M3 & Paratera/Tencent TokenHub & 2026-09-09--2026-10-22 & Output cap: 32,768 (2,762); 8,192 (1) \\
    \bottomrule
  \end{tabularx}
\end{table*}

\subsection{Output Classification and Geometric Scoring}
\label{sup:c2}

\textbf{Strict accuracy and content accuracy.} Every response is scored in two ways. Strict accuracy, the primary measure, counts a response as correct only if it both follows the output contract and gives the right answer; invalid responses, and responses whose format could merely be repaired, count as incorrect within the fixed denominator. The diagnostic content score instead asks whether the answer itself is right, and so first applies deterministic format recovery. The parser may unwrap a single-object array, strip a Markdown code fence, or extract a JSON object from surrounding text; for geometric answers, it may regroup a flat list of numbers into coordinate pairs, fix bracket errors in a point list, or evaluate an explicitly written arithmetic expression. A recovered response is scored against the same GT only if exactly one distinct answer satisfies the original contract. Recovery never infers missing fields and never changes the selected option, target image, or predicted geometry.

\label{sup:c2-chance-score}
\textbf{Chance-corrected score.} Judge and Select have different chance levels, so raw accuracies are not directly comparable. For raw strict accuracy $A$, we therefore report
\begin{equation}
  S=100\frac{A-b}{1-b}.
\end{equation}
Judge questions, in both the main benchmark and the SFT test, are binary ($b=0.50$), and all Select questions have four options ($b=0.25$). Uniform random answering thus scores 0 and a perfect answer set 100. Negative scores are kept, and invalid or missing answers count as incorrect within the fixed denominator. Region boxes, section lines, and adapted points are reported as raw accuracy (\%), and we never aggregate a total score across tasks or sources.

\label{sup:c2-output-contract}
\textbf{Output contract for the main benchmark.} The input contains the full question stem, any necessary floor or elevation information, and the drawings in image-index order. Judge/Select questions add lettered options in a fixed order, and Locate questions add the size of each original image and the index of the target image. The model must return a single JSON object, with no explanation or Markdown (\cref{tab:C2-1}).

\begin{table*}[t]
  \centering
  \footnotesize
  \setlength{\tabcolsep}{3pt}
  \caption{Output contract of the main benchmark: JSON format examples and their required meanings.}
  \label{tab:C2-1}
  \begin{tabularx}{\textwidth}{@{}>{\raggedright\arraybackslash}p{0.1\textwidth}>{\raggedright\arraybackslash}p{0.5\textwidth}L@{}}
    \toprule
    Format & JSON format example & Required meaning \\
    \midrule
    Judge/Select & \texttt{\{"answer":"A"\}} & The letter corresponds to the actual option order of the question; in this evaluation, Judge is binary and Select is four-option \\
    Region-box Locate & \texttt{\{"type":"region","image\_index":2,}\newline\texttt{"points":[[100,100],[200,100],[200,200],[100,200]]\}} & Specifies the image index; the four points give the four corners of the rectangle, starting from the top-left corner and proceeding clockwise along the boundary \\
    Section-line Locate & \texttt{\{"type":"line","image\_index":2,}\newline\texttt{"points":[[100,100],[100,200]],"direction":"right"\}} & Specifies the image index and two endpoints; the direction is one of \texttt{up/down/left/right}, pointing from the line toward the viewed side in the target-image canvas \\
    \bottomrule
  \end{tabularx}
\end{table*}

Coordinates are part of the task. Models output pixel coordinates in the given original image size, with the origin at the top-left corner, $x$ pointing right, $y$ pointing down, and $0\leq x\leq W$, $0\leq y\leq H$. Scoring maps them to the normalized canvas $(x/W,y/H)$. The human interface maps displayed positions onto the same canvas, so humans and models are judged by the same final GT and rules. The fine-tuning study in Sec.~6 of the main paper instead writes coordinates in a normalized 0--1000 range, as described in \cref{sup:g}.

\textbf{Geometric pass criteria.} A predicted region box $B$ passes if it overlaps sufficiently with one of the final acceptable boxes $\mathcal G$ of the question:
\begin{equation}
  \max_{G\in\mathcal G}\operatorname{IoU}(B,G)
  =\max_{G\in\mathcal G}\frac{|B\cap G|}{|B\cup G|}\geq0.60.
\end{equation}

\label{sup:c2-line-band}
Section lines are scored against the band adjudicated in \cref{sup:b1-line-band}: a single accepted positional band plus a viewing direction. The two boundaries are parallel in the pixel space of the original image, and every position between them, boundaries included, is acceptable. We write the boundaries as $n\cdot x=l$ and $n\cdot x=u$, with $n$ the unit normal. The GT also stores an evaluation span $t_1,t_2$ along the band, which fixes the horizontal or vertical interval where the check is made. If $z_1,z_2$ are the normal coordinates of the predicted supporting line at these two positions, the line passes geometrically when
\begin{equation}
  l\leq z_1\leq u,\qquad l\leq z_2\leq u.
\end{equation}

\begin{figure*}[t]
  \centering
  \includegraphics[width=\textwidth]{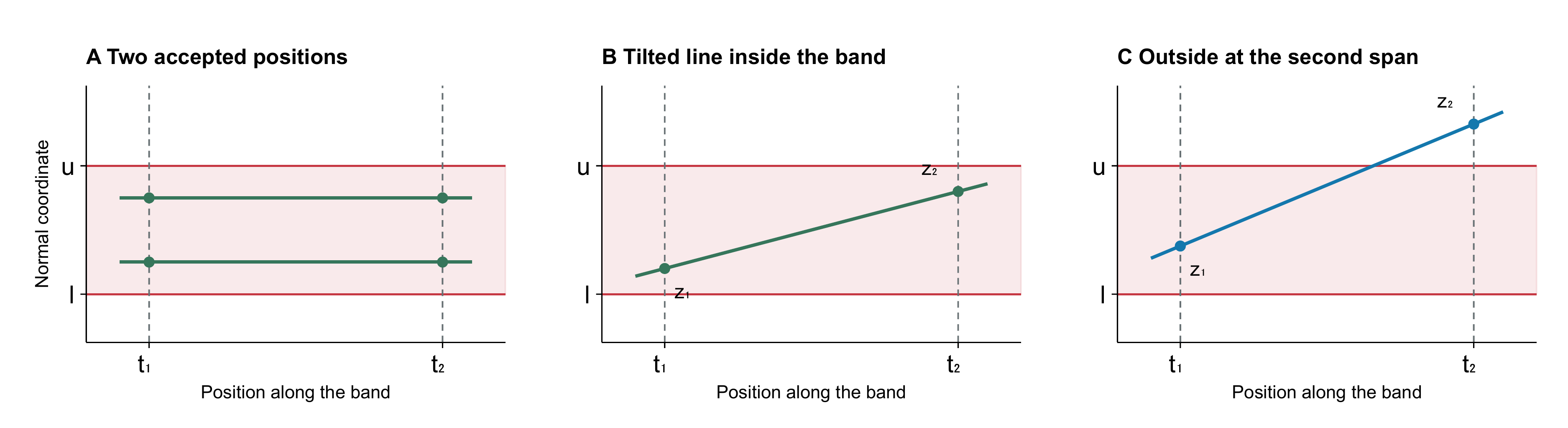}
  \caption{Band scoring schematic. The horizontal axis is the position along the band and the vertical axis is the normal coordinate; the shaded band has boundaries $l$ and $u$, and dashed lines mark the fixed evaluation span $t_1$ and $t_2$. A: two different line positions both lie within the range. B: a tilted line falls inside the band at both span positions. C: the prediction exits the band at $t_2$. Dots mark the intersections of the supporting line with the span positions. All three schematics assume a correct viewing direction.}
  \label{fig:S1a}
\end{figure*}

In addition, the viewing direction must match the GT. Checking both ends of the span constrains the cut position while tolerating a small error in the angle of the drawn line (\cref{fig:S1a}). Lines of zero length or perpendicular to the band axis fail. The representative line stored with each band is used only for single-line supervision, visualization, and distance and angle diagnostics.

A few conventions complete the region GT. Window annotations cover the window body and its own frame, and complete-space localization uses the minimum bounding rectangle of the target space. Each question names its target image and original image size.

Box localization accuracy is the share of questions that reach the IoU threshold, and \cref{sup:d5-threshold-sensitivity} varies this threshold. All results are reported separately by source, answer format, and geometry type, with questions weighted equally within each cell; stratum-weighted Human300 differences are given in \cref{sup:e1}.

\subsection{Statistical Units and Uncertainty}
\label{sup:c3}

\textbf{Samples and statistical units.} Model comparisons use the 2,763 stratified evaluation questions. Questions from the same building tend to be correlated, so we treat the building, not the question, as the resampling unit. Pair-level analyses use only the matched question pairs that meet the eligibility criteria of \cref{sup:d3}.

\label{sup:c3-statistical-support}
\textbf{Uncertainty and sensitivity to building composition.} The analysis covers 12 models, 194 matched-target question pairs, and the fixed Human300 answer sheets. Within each Natural/BIM $\times$ answer-format cell, we resample as many buildings as the cell contains, uniformly and with replacement, keep all questions of each drawn building, and repeat this 10,000 times. All models, and both sides of every paired comparison, share the same resamples. The 2.5th and 97.5th percentiles give 95\% reference intervals, which we report for the matched-target differences (\cref{tab:D3-1}) and the Human300 comparison (\cref{tab:S21}).

For Human300, the fixed analysis weights over the 41 strata are kept. In each replicate, a participant's accuracy is the weighted correct count divided by the weighted denominator; the two architects are averaged with equal weight, and their answers are resampled jointly with the model's on the same questions.

\textbf{Small pools.} Building resampling needs enough buildings to be stable. The Natural region pairs (134 pairs / 22 buildings) meet this need and receive intervals; Natural line (26/14), BIM region (25/7), and BIM line (9/6) do not, and are reported as counts only (\cref{tab:S17}).

\section{Full Evaluation and Sensitivity}
\label{sup:d}

This section expands Sec.~4 of the main paper. We first give the overall scores and the full task breakdown (\cref{sup:d1,sup:d2}), then describe how Select and Locate questions are paired by target (\cref{sup:d3}), and finally examine how scoring thresholds and re-annotation affect the results (\cref{sup:d5-threshold-sensitivity}).

\subsection{Overall, Region, and Content Scores}
\label{sup:d1}

\textbf{Closed-form answer-distribution reference.} Closed-form scores are best read against what trivial strategies achieve. For Natural, Judge/Select have 794/435 questions; uniform guessing yields 50\%/25\%, and always choosing the majority label yields 50.00\%/28.74\%. For BIM (441/447 questions), the majority-label references are 53.06\%/25.95\%. The majority labels are a tie between A and B for Natural Judge, A for Natural Select, B for BIM Judge, and C for BIM Select. Main-paper Table~2 gives per-model results for the 12 frontier models.

Format recovery can matter for individual models. After deterministic recovery, Gemini's Natural region localization rises from 39/393 to 59/393, whereas Grok's correct count does not change.

\textbf{Smaller open-weight models.} \Cref{tab:open-weight} gives raw accuracies and chance-corrected scores for the ten smaller open-weight models of main-paper Table~2, run from the public checkpoints Qwen3.5-9B; Qwen3-VL-4B-Instruct and Qwen3-VL-8B-Instruct; Qwen2.5-VL-7B-Instruct; InternVL3\_5-8B-HF and InternVL3\_5-14B-HF; Molmo2-4B and Molmo2-O-7B; pixtral-12b; and Mistral-Small-3.2-24B-Instruct-2506. Localization is essentially out of reach for this group: no model localizes more than one box or one line on either source. Scores far below chance, such as Pixtral-12B's, arise because most of its answers break the output contract and therefore count as incorrect.

\begin{table*}[t]
  \centering
  \footnotesize
  \setlength{\tabcolsep}{4pt}
  \caption{Smaller open-weight models on the 2,763 evaluation questions. Judge/Select give chance-corrected scores (points) with raw accuracy (\%) in parentheses; Box/Line give strict accuracy (\%). Denominators are given in the header.}
  \label{tab:open-weight}
  \begin{tabular}{@{}lrrrrrrrr@{}}
    \toprule
    & \multicolumn{4}{c}{Natural} & \multicolumn{4}{c@{}}{BIM} \\
    \cmidrule(lr){2-5}\cmidrule(l){6-9}
    Model & Judge /794 & Select /435 & Box /393 & Line /50 & Judge /441 & Select /447 & Box /163 & Line /40 \\
    \midrule
    Qwen3.5-9B & 21.91 (60.96) & 13.56 (35.17) & 0.25 & 2.00 & 13.38 (56.69) & 10.22 (32.66) & 0.00 & 0.00 \\
    Qwen3-VL-4B & 13.35 (56.68) & 10.19 (32.64) & 0.00 & 0.00 & 13.38 (56.69) & 8.13 (31.10) & 0.61 & 0.00 \\
    Qwen3-VL-8B & 13.35 (56.68) & 8.05 (31.03) & 0.00 & 0.00 & 9.75 (54.88) & 5.15 (28.86) & 0.00 & 0.00 \\
    Qwen2.5-VL-7B & 15.11 (57.56) & 6.51 (29.89) & 0.00 & 2.00 & 1.13 (50.57) & 5.44 (29.08) & 0.61 & 0.00 \\
    InternVL3.5-8B & 1.51 (50.76) & 4.37 (28.28) & 0.25 & 0.00 & 7.48 (53.74) & 4.55 (28.41) & 0.00 & 0.00 \\
    InternVL3.5-14B & 5.04 (52.52) & 5.90 (29.43) & 0.25 & 2.00 & 7.03 (53.51) & 12.60 (34.45) & 0.00 & 0.00 \\
    Molmo2-4B & 2.27 (51.13) & 1.00 (25.75) & 0.00 & 0.00 & 7.94 (53.97) & 21.25 (40.94) & 0.61 & 0.00 \\
    Molmo2-O-7B & 2.77 (51.39) & 1.00 (25.75) & 0.00 & 0.00 & 11.56 (55.78) & 16.78 (37.58) & 0.61 & 0.00 \\
    Pixtral-12B & $-$89.42 (5.29) & $-$30.57 (2.07) & 0.00 & 0.00 & $-$46.94 (26.53) & $-$14.54 (14.09) & 0.00 & 0.00 \\
    Mistral Small 3.2-24B & 2.02 (51.01) & 5.90 (29.43) & 0.00 & 0.00 & $-$2.04 (48.98) & 5.15 (28.86) & 0.61 & 0.00 \\
    \bottomrule
  \end{tabular}
\end{table*}

\subsection{Full Task Breakdown}
\label{sup:d2}

\begin{figure*}[t]
  \centering
  \includegraphics[width=\textwidth]{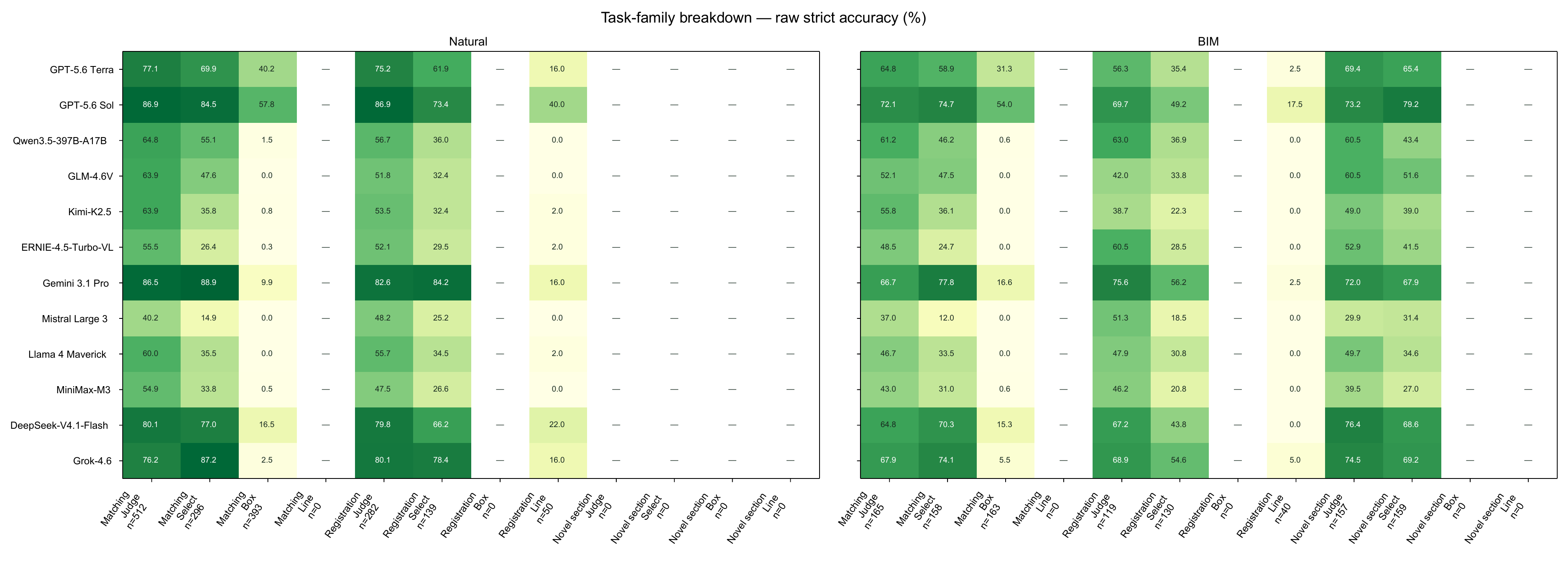}
  \caption{Raw strict accuracy by task category and answer format. Both panels share a 0--100\% color scale; missing cells indicate tasks not covered by that source. Closed-form scores in the main paper are chance-corrected as described in \cref{sup:c2}; matched-target comparisons appear in \cref{sup:d3}.}
  \label{fig:S2}
\end{figure*}

Aggregate scores hide where models succeed. The task breakdown (\cref{fig:S2}) shows that placing a directed line is far harder than the other questions of the same category. In Projection Registration on Natural, Sol answers 245/282 Judge and 102/139 Select questions correctly (86.9\% and 73.4\% raw accuracy) but only 20/50 directed-line questions (40.0\%). On BIM the pattern repeats, with 83/119, 64/130, and 7/40 (69.7\%, 49.2\%, and 17.5\%).

Model rankings also shift from task to task. On BIM Cross-view Matching Select, Gemini (123/158, 77.8\%) edges out Sol (118/158, 74.7\%), yet on BIM Novel-section Prediction Select, Sol (126/159, 79.2\%) is well ahead of Gemini (108/159, 67.9\%). These are raw accuracies; the main text reports the corresponding chance-corrected scores. Because the order changes with the spatial query, no single aggregate score describes a model's performance adequately.

\subsection{Pairing by Target and Eligibility}
\label{sup:d3}

The Select--Locate comparison of the main paper relies on question pairs that ask about the same target in two formats. Paired questions share the target identity, the reference drawing, and the target base drawing, and differ only in that Select offers candidates whereas Locate requires free-form output. The common pool requires gradable answers on both sides for all 12 models and, within each model, the same model identity, service provider, explicit inference parameters, output limit, and response mode.

\textbf{Filtering procedure.} Of the 285 candidate pairs matched by target, 91 fail these criteria. The remaining 194 pairs span 32 buildings: 134 region and 26 line pairs from Natural, and 25 region and 9 line pairs from BIM. All Select--Locate differences we report refer to these pairs or their subsets.

\Cref{tab:S17} gives the contingency counts for all 12 models, and \cref{tab:D3-1} adds whole-building bootstrap intervals for Natural regions.

\begin{table}[t]
  \centering
  \scriptsize
  \setlength{\tabcolsep}{2pt}
    \caption{Matched-target content scoring. Each cell lists, in order, both correct / Select-only correct / Locate-only correct / both incorrect.}
    \label{tab:S17}
    \begin{tabular}{@{}lcccc@{}}
      \toprule
      & \multicolumn{2}{c}{Natural} & \multicolumn{2}{c}{BIM} \\
      \cmidrule(lr){2-3}\cmidrule(l){4-5}
      Model & Region /134 & Line /26 & Region /25 & Line /9 \\
      \midrule
    Terra & 43/53/2/36 & 4/14/0/8 & 9/8/1/7 & 0/2/0/7 \\
    Sol & 63/48/5/18 & 10/8/1/7 & 15/6/0/4 & 0/4/0/5 \\
    Gemini & 16/104/0/14 & 4/13/0/9 & 4/16/0/5 & 0/3/0/6 \\
    Grok-4.6 & 0/119/0/15 & 4/16/2/4 & 0/21/1/3 & 0/3/1/5 \\
    DeepSeek & 16/91/1/26 & 7/10/1/8 & 3/17/0/5 & 0/2/0/7 \\
    Qwen3.5-397B & 0/81/0/53 & 0/14/0/12 & 0/11/0/14 & 0/4/0/5 \\
    GLM-4.6V & 0/67/0/67 & 0/9/0/17 & 0/12/0/13 & 0/5/0/4 \\
    Kimi-K2.5 & 3/64/0/67 & 0/15/0/11 & 0/15/0/10 & 0/1/0/8 \\
    ERNIE & 0/37/0/97 & 0/7/0/19 & 0/4/0/21 & 0/4/0/5 \\
    Mistral & 0/35/0/99 & 0/8/0/18 & 0/4/0/21 & 0/4/0/5 \\
    Llama & 0/42/0/92 & 0/9/1/16 & 0/8/0/17 & 0/4/0/5 \\
    MiniMax & 1/56/0/77 & 0/9/0/17 & 1/12/0/12 & 0/3/0/6 \\
    \bottomrule
    \end{tabular}
\end{table}

The asymmetry is stark on the 134 Natural region pairs. Under content scoring, Sol is right on Select but wrong on Locate in 48 pairs and the reverse in only 5; for Gemini the counts are 104 and 0, and for Grok 119 and 0. \Cref{tab:D3-1} turns these counts into Select$-$Locate differences with building-level intervals.

\begin{table}[t]
  \centering
  \footnotesize
  \setlength{\tabcolsep}{3pt}
  \caption{Raw-accuracy Select$-$Locate differences under content scoring on the 134 Natural matched-target region pairs for the models discussed in the main text, with 95\% whole-building bootstrap intervals.}
  \label{tab:D3-1}
  \begin{tabular}{@{}lcc@{}}
    \toprule
    Model & Select$-$Locate (pp) & 95\% bootstrap interval \\
    \midrule
    Sol & 32.09 & [20.83, 45.54] \\
    Gemini & 77.61 & [68.15, 86.92] \\
    Grok-4.6 & 88.81 & [80.15, 96.15] \\
    \bottomrule
  \end{tabular}
\end{table}

\subsection{Sensitivity to Thresholds and Re-annotation}
\label{sup:d5-threshold-sensitivity}
\label{sup:d4}

\textbf{Scoring thresholds.} 
We next vary the scoring thresholds while holding answers and GT fixed. For regions, we use IoU thresholds of 0.50, 0.60, and 0.70. For lines, the adjudicated band gives the primary result, and widening each boundary outward by 1 original-image pixel tests sensitivity; both conditions still require the correct viewing direction.

Sol's lead does not depend on the threshold: it is the best of the 12 models in region localization at every threshold on both sources, with Natural accuracies of 66.67\%, 57.76\%, and 47.84\% (393 questions) and BIM accuracies of 57.67\%, 53.99\%, and 49.69\% (163 questions). The Select--Locate gap of Gemini and Grok on matched region pairs also persists at the most lenient threshold (\cref{tab:D5-1}).

\begin{table}[t]
  \centering
  \footnotesize
  \setlength{\tabcolsep}{3pt}
  \caption{Raw strict Select and Locate accuracy under three IoU thresholds on the fixed matched-target region pairs for Gemini and Grok.}
  \label{tab:D5-1}
  \begin{tabular}{@{}lrrrrr@{}}
    \toprule
    & & & \multicolumn{3}{c}{Locate} \\
    \cmidrule(l){4-6}
    Model/source & Pairs & Select & IoU$\geq$0.50 & IoU$\geq$0.60 & IoU$\geq$0.70 \\
    \midrule
    Gemini/Natural & 134 & 89.55\% & 13.43\% & 7.46\% & 4.48\% \\
    Gemini/BIM & 25 & 80.00\% & 20.00\% & 16.00\% & 12.00\% \\
    Grok/Natural & 134 & 88.81\% & 1.49\% & 0.00\% & 0.00\% \\
    Grok/BIM & 25 & 84.00\% & 4.00\% & 4.00\% & 4.00\% \\
    \bottomrule
  \end{tabular}
\end{table}

Widening the line band by 1 pixel changes no verdict for the 12 models, either on the main set or on the matched Human300 items; only Architect B gains one BIM item.

\label{sup:d5-line-band-audit}
\textbf{Band width and out-of-band distance.} The accepted bands are narrow. The median band covers 2.01\% (Natural) and 1.98\% (BIM) of the image extent normal to it, and the two-end check permits a median line tilt of only 1.43$^\circ$ and 1.07$^\circ$. All 69 strictly correct model answers are parallel to the band axis. Failed lines with the correct viewing direction, in contrast, miss by a wide margin: they lie a median of 176.35 (Natural) and 385.50 (BIM) original-image pixels outside the band, and only four per source fall within 5 pixels of a boundary.

\label{sup:d5-line-band-reannotation}
\textbf{Independent re-annotation.} Finally, we check how much the results depend on which expert drew the band. At the two evaluation positions used in scoring, we compare the two annotations' intervals along the normal of the first band and average the two interval IoUs to obtain a per-question overlap. The boundary deviation is the larger offset of the two side boundaries over the two positions. For re-scoring, we substitute the second expert's boundaries and direction while keeping the evaluation span, model answers, and scoring rules unchanged (\cref{tab:S20c}).

\begin{table}[t]
  \centering
  \footnotesize
  \setlength{\tabcolsep}{3pt}
  \caption{Agreement between the two independently annotated positional bands and re-scoring results. Arrows point from the first (primary) scoring to scoring under the second annotation.}
  \label{tab:S20c}
  \begin{tabularx}{\columnwidth}{@{}Lrr@{}}
    \toprule
    Metric & Natural & BIM \\
    \midrule
    Same viewing direction & 50/50 & 40/40 \\
    Overlap at both evaluation positions & 50/50 & 40/40 \\
    Median per-question interval IoU & 0.897 & 0.893 \\
    Median per-question maximum boundary deviation (orig.\ px) & 3.43 & 4.42 \\
    Model answers with changed verdict & 4/600 & 1/480 \\
    Sol main-evaluation line correct & 20/50 $\rightarrow$ 19/50 & 7/40 $\rightarrow$ 7/40 \\
    Human300 architect mean minus Sol (pp) & +51.92 $\rightarrow$ +53.85 & +67.39 $\rightarrow$ +67.39 \\
    \bottomrule
  \end{tabularx}
\end{table}

The two annotations agree closely, and verdicts barely move. Five of the 1,080 model verdicts change, on five questions: four become correct and one incorrect. Three of the 98 architect answers change.

\section{Human Study}
\label{sup:e}

This section describes how Human300 was collected and weighted, and compares the architects with a fixed model comparator on the same items.

\subsection{Human300 Collection, Weights, and Matched-Item Results}
\label{sup:e1}

\textbf{Human300 collection.} Two practicing architects, each with about 15 years of professional experience, independently answered the same 300 questions (150 Natural, 150 BIM) under blind conditions. They saw exactly what the models saw, with identical drawings and candidate options, and were scored against the same GT and criteria. As the comparator we fix Sol, which leads directed-line localization on both sources in the full evaluation (\cref{tab:S21}). Quality review is described in \cref{sup:b2}.

\textbf{Informed consent.} Both architects provided informed consent to participate and to the use of their responses for research.

\label{sup:e1-sampling-weights}
\textbf{Fixed stratification and weights.} Human300 samples 300 questions by simple random sampling without replacement from 41 strata, defined by source $\times$ task subtype $\times$ answer format, under predefined quotas and seed 20261016, and every participant answers the same list. The sampling frame is the 2,765-question pool of \cref{sup:b1}, so the two Natural questions removed afterward were not sampled. Inclusion probabilities are computed on this frame. Because the quotas do not follow the composition of the evaluation set, each sampled question carries the analysis weight $w_h=N_h/n_h$, where $N_h$ is the number of the 2,763 evaluation questions in stratum $h$ and $n_h$ is the number of sampled questions. \Cref{tab:E1-1} lists these weights.

\begin{table}[t]
  \centering
  \footnotesize
  \setlength{\tabcolsep}{3pt}
  \caption{Human300 stratum weights $N_h/n_h$ by source, task subtype, and answer format. ``---'' indicates no corresponding stratum.}
  \label{tab:E1-1}
  \begin{tabularx}{\columnwidth}{@{}lLrrrr@{}}
    \toprule
    Source & Task subtype & Judge & Select & Box & Line \\
    \midrule
    Natural & Component--component & 215/13 & 215/14 & 214/25 & --- \\
    Natural & Space--space & 82/5 & 14/2 & 49/6 & --- \\
    Natural & Multi-view correspondence & 215/14 & 67/4 & 130/15 & --- \\
    Natural & Section-to-plan line & 100/6 & 48/3 & --- & 50/26 \\
    Natural & Elevation direction & 182/11 & 91/6 & --- & --- \\
    \midrule
    BIM & Component--component & 40/4 & 40/4 & 42/10 & --- \\
    BIM & Space--space & 40/4 & 40/4 & 40/10 & --- \\
    BIM & Component--space & 44/4 & 37/4 & 38/9 & --- \\
    BIM & Multi-view correspondence & 41/4 & 41/4 & 43/11 & --- \\
    BIM & Section-to-plan line & 41/4 & 45/4 & --- & 40/23 \\
    BIM & Elevation direction & 37/4 & 42/4 & --- & --- \\
    BIM & Relative section registration & 41/4 & 43/4 & --- & --- \\
    BIM & Whole-section selection & 41/4 & 43/4 & --- & --- \\
    BIM & Object visibility & 38/4 & 39/4 & --- & --- \\
    BIM & Object order & 35/3 & 41/4 & --- & --- \\
    BIM & Object count/attribute & 43/4 & 36/4 & --- & --- \\
    \bottomrule
  \end{tabularx}
\end{table}

Within each source/answer-format cell $c$, let $y_{ri}$ be the 0/1 score of participant or model $r$ on question $i$, determined from the final adjudicated answer and the scoring rule for that format. Then
\begin{equation}
  \begin{aligned}
    \widehat A_{r,c}&=\frac{\sum_{h\in c}w_h\sum_{i\in S_h}y_{ri}}{\sum_{h\in c}N_h},\\
    \Delta_c&=100\left(\frac{\widehat A_{A,c}+\widehat A_{B,c}}{2}-\widehat A_{\mathrm{Sol},c}\right).
  \end{aligned}
\end{equation}
Here $S_h$ is the set of sampled questions in stratum $h$, $A,B$ are the two architects, and $\Delta_c$ is in percentage points. The weights thus restore the question composition of the evaluation set within each source and answer format.

\begin{table*}[t]
  \centering
  \footnotesize
  \setlength{\tabcolsep}{3pt}
  \caption{Matched-item comparison on Human300. Participant columns report sample correct counts and accuracy (closed-form results not chance-corrected); the difference is the architect mean minus Sol, weighted over the 41 sampling strata. Boxes use IoU $\geq$ 0.60; lines use the viewing direction and the accepted positional band. The intervals and leave-one-building-out analysis measure sensitivity to building composition under fixed participants and weights; the method is described in \cref{sup:c3}.}
  \label{tab:S21}
  \begin{tabularx}{\textwidth}{@{}llrrrr>{\raggedleft\arraybackslash}X>{\raggedleft\arraybackslash}X>{\raggedleft\arraybackslash}X@{}}
    \toprule
    Source & Answer format & Per person & Architect A & Architect B & Sol & Weighted architect mean$-$Sol (pp) & 95\% reference interval of difference (pp) & Leave-one-building-out range (pp) \\
    \midrule
    Natural & Judge & 49 & 49/49 (100.0\%) & 48/49 (98.0\%) & 45/49 (91.8\%) & +7.00 & [0.96, 14.74] & [5.01, 7.64] \\
    Natural & Select & 29 & 28/29 (96.6\%) & 29/29 (100.0\%) & 21/29 (72.4\%) & +26.30 & [12.44, 43.09] & [20.74, 28.35] \\
    Natural & Region box & 46 & 44/46 (95.7\%) & 46/46 (100.0\%) & 24/46 (52.2\%) & +45.76 & [33.25, 58.01] & [44.30, 48.87] \\
    Natural & Section line + direction & 26 & 24/26 (92.3\%) & 23/26 (88.5\%) & 10/26 (38.5\%) & +51.92 & [29.41, 68.33] & [45.24, 56.25] \\
    \midrule
    BIM & Judge & 43 & 41/43 (95.3\%) & 42/43 (97.7\%) & 33/43 (76.7\%) & +19.43 & [2.42, 36.93] & [13.44, 24.82] \\
    BIM & Select & 44 & 43/44 (97.7\%) & 42/44 (95.5\%) & 35/44 (79.5\%) & +16.83 & [6.59, 29.60] & [13.64, 20.79] \\
    BIM & Region box & 40 & 38/40 (95.0\%) & 37/40 (92.5\%) & 22/40 (55.0\%) & +39.19 & [25.51, 49.40] & [36.49, 42.68] \\
    BIM & Section line + direction & 23 & 20/23 (87.0\%) & 21/23 (91.3\%) & 5/23 (21.7\%) & +67.39 & [40.38, 92.12] & [64.29, 80.00] \\
    \bottomrule
  \end{tabularx}
\end{table*}

\section{Behavioral Diagnostics}
\label{sup:f}

The diagnostics in this section probe why cross-view localization fails. All 12 models answer the 495 basic-reading and 200 geometric-reproduction questions; reference switching covers five models with 248 conditions each. Each diagnostic is scored against its own GT.

\subsection{Basic Drawing Reading and Geometric Output Control}
\label{sup:f1}

A low cross-view score could also mean that a model cannot read the drawings or cannot write coordinates. The single-image controls separate these possibilities with 495 reading questions and 200 geometric reproduction questions, 695 in total (399 Natural, 296 BIM), giving 8,340 responses from the 12 models. Reading scores and geometric accuracy are reported separately.

\subsubsection{Basic Drawing Reading}
\label{sup:f1-reading}

The 495 reading questions follow the source, category, and drawing-coverage distribution of the main set: 299 Natural and 196 BIM, covering 16 categories including object categories, symbols/text, and artificial marks. Each is a four-choice question that asks the model to name the component inside a mark, read a symbol's direction or a piece of text, or match a colored box or region to its number; no coordinates are required, and all 12 models return complete responses.

The Natural denominators for object, symbol/text, and mark questions are 227, 49, and 23, and the BIM denominators are 139, 42, and 15. As in \cref{sup:c2}, four-choice scores are chance-corrected (\cref{tab:S24a}). Sol and Gemini score at least 90.41 points in every category, so their cross-view failures cannot be attributed to basic reading.

\begin{table*}[t]
  \centering
  \footnotesize
  \setlength{\tabcolsep}{3pt}
  \caption{Basic-reading scores (points) of the 12 models. N/B denote Natural/BIM.}
  \label{tab:S24a}
  \begin{tabularx}{\textwidth}{@{}l*{6}{>{\raggedleft\arraybackslash}X}@{}}
    \toprule
    Model & N object & N symbol/text & N mark & B object & B symbol/text & B mark \\
    \midrule
    Terra & 94.13 & 100.00 & 82.61 & 83.69 & 90.48 & 91.11 \\
    Sol & 95.89 & 100.00 & 94.20 & 90.41 & 90.48 & 100.00 \\
    Gemini & 99.41 & 100.00 & 100.00 & 90.41 & 90.48 & 100.00 \\
    Grok-4.6 & 94.71 & 97.28 & 88.41 & 88.49 & 96.83 & 100.00 \\
    DeepSeek & 92.95 & 100.00 & 94.20 & 87.53 & 93.65 & 100.00 \\
    Qwen3.5-397B & 88.84 & 100.00 & 88.41 & 80.82 & 96.83 & 100.00 \\
    GLM-4.6V & 79.44 & 91.84 & 88.41 & 60.67 & 71.43 & 100.00 \\
    Kimi-K2.5 & 77.68 & 91.84 & 71.01 & 70.26 & 84.13 & 91.11 \\
    ERNIE & 74.74 & 59.18 & 18.84 & 58.75 & 39.68 & 37.78 \\
    Mistral & 33.04 & 78.23 & 76.81 & 24.22 & 61.90 & 100.00 \\
    Llama & 63.58 & 89.12 & 71.01 & 56.83 & 65.08 & 100.00 \\
    MiniMax & 80.03 & 86.39 & 36.23 & 68.35 & 61.90 & 82.22 \\
    \bottomrule
  \end{tabularx}
\end{table*}

\subsubsection{Single-image Geometric Reproduction and Matched-target Cross-view Comparison}
\label{sup:f1-geometric-controls}

\textbf{Questions and sampling.} The reproduction controls test whether a model can copy a mark it is shown. They contain 100 box and 100 directed-line questions, each split evenly between Natural and BIM. Boxes are selected deterministically from the main evaluation to cover buildings, quartiles of target area, and the original subtypes, using only questions with a single GT box. The line questions comprise all 90 main-evaluation line questions plus 10 BIM controls from the same buildings (GNI 6, LR 4); these 10 serve only the single-image controls and enter neither the main evaluation nor the 190-question paired set.

\textbf{Inputs, GT, and scoring.} Each control draws a red rectangle, or a red section symbol with its viewing direction, directly on a single full-size base drawing, and the model must reproduce that marking; the drawing is not cropped. The centerline of the drawn box coincides with the adopted GT, without component dilation, and lines are drawn in full with end caps and short direction ticks. Answers use the pixel coordinates of the main evaluation: four corners for boxes, and two endpoints plus a viewing direction for lines. Each of the 190 paired questions uses the same GT and scoring as its cross-view counterpart, namely box IoU $\geq$ 0.60, or the correct direction and placement within the accepted band, and the red line of every paired line question lies inside that band. The other 10 BIM line controls are scored by distance to the marked line.

\begin{figure}[t]
  \centering
  \includegraphics[width=\columnwidth]{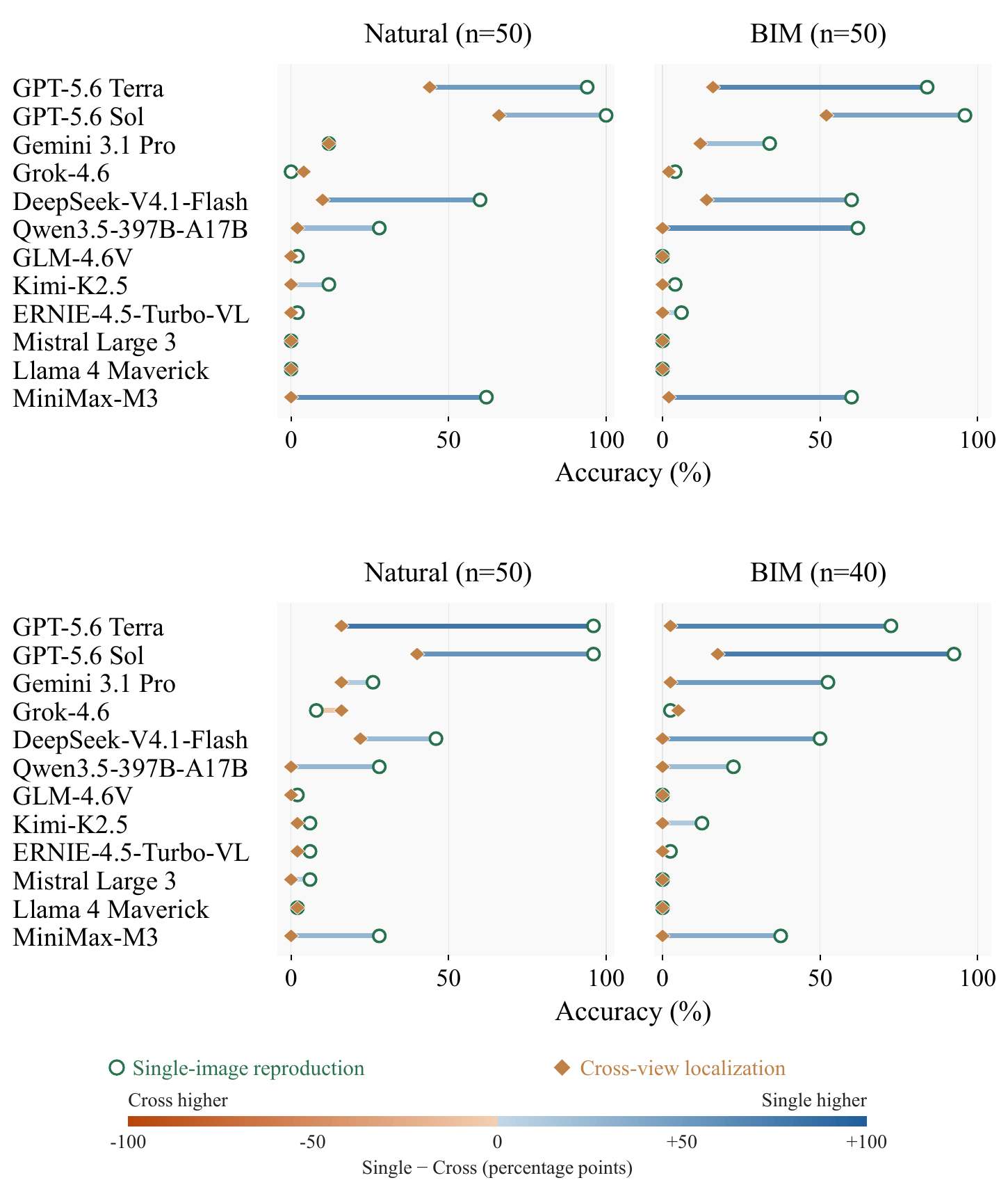}
  \caption{Matched single-image reproduction (circles) and cross-view localization (diamonds) on 100 box questions (top) and 90 directed-line questions (bottom), by source. Connector hue and intensity encode the direction and size of the gap.}
  \label{fig:reproduction}
\end{figure}

\textbf{Model settings.} All 12 models answer the 200 controls, and the 190 paired questions use the same output limits and explicit reasoning settings as their main-evaluation counterparts, with images at original size. The contrast is large: pooled over the 12 models, 704 of the 2,280 matched targets are reproduced, but only 187 are localized across views. \Cref{fig:reproduction} compares the two conditions model by model, and \cref{tab:S24d} gives the paired outcomes.

\begin{table*}[t]
  \centering
  \footnotesize
  \setlength{\tabcolsep}{3pt}
  \caption{Strict scoring of the 190 matched-target questions. Each cell in the first four columns lists, in order: both correct, only single-image reproduction correct, only cross-view localization correct, and both incorrect. The last column gives cross-view correct/questions and accuracy within the reproduction-success subset.}
  \label{tab:S24d}
  \begin{tabularx}{\textwidth}{@{}l*{4}{>{\centering\arraybackslash}X}r@{}}
    \toprule
    Model & Natural box /50 & BIM box /50 & Natural line /50 & BIM line /40 & Cross-view acc.\ on reproduction-success subset \\
    \midrule
    Terra & 22/25/0/3 & 8/34/0/8 & 8/40/0/2 & 0/29/1/10 & 38/166 (22.9\%) \\
    Sol & 33/17/0/0 & 25/23/1/1 & 20/28/0/2 & 7/30/0/3 & 85/183 (46.4\%) \\
    Gemini & 1/5/5/39 & 3/14/3/30 & 3/10/5/32 & 1/20/0/19 & 8/57 (14.0\%) \\
    Grok-4.6 & 0/0/2/48 & 1/1/0/48 & 1/3/7/39 & 0/1/2/37 & 2/7 (28.6\%) \\
    DeepSeek & 5/25/0/20 & 6/24/1/19 & 9/14/2/25 & 0/20/0/20 & 20/103 (19.4\%) \\
    Qwen3.5-397B & 1/13/0/36 & 0/31/0/19 & 0/14/0/36 & 0/9/0/31 & 1/68 (1.5\%) \\
    GLM-4.6V & 0/1/0/49 & 0/0/0/50 & 0/1/0/49 & 0/0/0/40 & 0/2 (0.0\%) \\
    Kimi-K2.5 & 0/6/0/44 & 0/2/0/48 & 1/2/0/47 & 0/5/0/35 & 1/16 (6.2\%) \\
    ERNIE & 0/1/0/49 & 0/3/0/47 & 0/3/1/46 & 0/1/0/39 & 0/8 (0.0\%) \\
    Mistral & 0/0/0/50 & 0/0/0/50 & 0/3/0/47 & 0/0/0/40 & 0/3 (0.0\%) \\
    Llama & 0/0/0/50 & 0/0/0/50 & 0/1/1/48 & 0/0/0/40 & 0/1 (0.0\%) \\
    MiniMax & 0/31/0/19 & 1/29/0/20 & 0/14/0/36 & 0/15/0/25 & 1/90 (1.1\%) \\
    \bottomrule
  \end{tabularx}
\end{table*}

The last column of \cref{tab:S24d} sums ``both correct'' over the four source/geometry combinations and divides by the sum of ``both correct'' and ``only single-image reproduction correct.''

Even models that copy a line correctly rarely place it across views: of the BIM line questions they reproduce, Sol and Terra fail 30/37 and 29/29 in the cross-view condition. The difference lies in the input. The marked single image already supplies both the cut line and its direction, whereas the cross-view question requires inferring them from the section.

\label{sup:f1-coordinate-sensitivity}
\textbf{Coordinate-unit sensitivity.} A model might also write coordinates in a unit other than pixels. To see how much this affects scores, we reinterpret each model's coordinates as original-image pixels, 0--1000 values, or 0--1 values, applying one interpretation to all of its responses with the GT fixed. For each model, we pick the unit that gives the most correct answers on the 190 single-image controls (ties favor pixels) and apply it to that model's cross-view responses. GLM, Gemini, and Llama select 0--1000; all other models select pixels. \Cref{tab:control-selected-units} reports the resulting accuracies; the main evaluation keeps the unified pixel contract of \cref{sup:c2}.

\begin{table*}[t]
  \centering
  \footnotesize
  \setlength{\tabcolsep}{4pt}
  \caption{Control-selected coordinate units for the localization columns of the 12 frontier models in main-paper Table~2. Each cell reports strict accuracy (\%) under the original pixel contract $\rightarrow$ the control-selected unit; a single value denotes unchanged accuracy. One unit per model is selected using only the 190 single-image controls and applied to both sources and geometries. Judge/Select scores are unchanged.}
  \label{tab:control-selected-units}
  \begin{tabularx}{\textwidth}{@{}ll*{4}{>{\centering\arraybackslash}X}@{}}
    \toprule
     & & \multicolumn{2}{c}{Natural} & \multicolumn{2}{c}{BIM} \\
    \cmidrule(lr){3-4}\cmidrule(lr){5-6}
    Model & Selected unit & Box /393 & Line /50 & Box /163 & Line /40 \\
    \midrule
    Terra & Pixels & 40.2 & 16.0 & 31.3 & 2.5 \\
    Sol & Pixels & 57.8 & 40.0 & 54.0 & 17.5 \\
    Gemini & 0--1000 & 9.9 $\rightarrow$ 5.1 & 16.0 $\rightarrow$ 6.0 & 16.6 $\rightarrow$ 3.7 & 2.5 $\rightarrow$ 0.0 \\
    Grok-4.6 & Pixels & 2.5 & 16.0 & 5.5 & 5.0 \\
    DeepSeek & Pixels & 16.5 & 22.0 & 15.3 & 0.0 \\
    Qwen3.5-397B & Pixels & 1.5 & 0.0 & 0.6 & 0.0 \\
    GLM-4.6V & 0--1000 & 0.0 $\rightarrow$ 14.5 & 0.0 $\rightarrow$ 2.0 & 0.0 $\rightarrow$ 6.1 & 0.0 \\
    Kimi-K2.5 & Pixels & 0.8 & 2.0 & 0.0 & 0.0 \\
    ERNIE & Pixels & 0.3 & 2.0 & 0.0 & 0.0 \\
    Mistral & Pixels & 0.0 & 0.0 & 0.0 & 0.0 \\
    Llama & 0--1000 & 0.0 $\rightarrow$ 0.3 & 2.0 & 0.0 $\rightarrow$ 0.6 & 0.0 \\
    MiniMax & Pixels & 0.5 & 0.0 & 0.6 & 0.0 \\
    \bottomrule
  \end{tabularx}
\end{table*}

The conversion helps GLM most, raising its Natural/BIM box accuracy to 14.5\%/6.1\%, while Llama gains only one box per source. Gemini's cross-view box and line accuracies actually decline on both sources, even though its single-image reproduction improves. The other nine models keep pixels and unchanged scores. Unit choice thus shifts the scores of a few models, but cross-view localization stays low under the control-selected units.

\subsubsection{Ablation of Reference Information}
\label{sup:f1-reference-ablation}

\textbf{Purpose and sampling.} If a model truly matches across views, removing the reference should hurt it. We test this on 50 four-choice component correspondence questions sampled from the fixed evaluation set, 25 Natural and 25 BIM, rotating across buildings so that they span 18 and 15 buildings, respectively. We report paired results for Terra and Sol.

\textbf{Input conditions.} Each question is answered twice: once with the full input, and once with the reference drawing replaced by a pure-white RGB image of the same size. Everything else stays identical, including the target drawing's pixels, image order, labels, question stem, candidates, and answer contract, so the two conditions differ only in the visual information from the reference. The two conditions are answered independently, in an order that alternates across questions.

\textbf{Settings and scoring.} Both models use high reasoning effort and identical configurations in both conditions, giving 200 responses. Both conditions are scored against the GT of the complete question, so the blanked condition measures how often a model still reaches the original answer without knowing the reference object. We report the four-choice chance-corrected score $S=100(A-0.25)/0.75$, under which random selection scores 0.

\begin{table}[t]
  \centering
  \footnotesize
  \setlength{\tabcolsep}{3pt}
  \caption{Reference-information ablation on 50 component-correspondence questions. The denominator is 25 for each model, source, and condition; parentheses give strict correct/questions, and scores are converted from unrounded accuracy. The expected chance-corrected score of Random is 0 points. All 200 responses comply with the output contract.}
  \label{tab:S24f}
  \begin{tabularx}{\columnwidth}{@{}ll*{2}{>{\raggedleft\arraybackslash}X}@{}}
    \toprule
    Model & Source & Full input: score (correct) & Blanked reference: score (correct) \\
    \midrule
    Terra & Natural & 46.7 (15/25) & 4.0 (7/25) \\
    Terra & BIM & 14.7 (9/25) & $-$6.7 (5/25) \\
    Sol & Natural & 78.7 (21/25) & 4.0 (7/25) \\
    Sol & BIM & 52.0 (16/25) & 4.0 (7/25) \\
    \bottomrule
  \end{tabularx}
\end{table}

\textbf{Relation to reference switching.} Without the reference, Terra and Sol fall to 20\%--28\% raw accuracy across the two sources, close to the 25\% chance level, compared with 36\%--84\% with the full input. Both models therefore use the reference on these questions. Using the reference is not the same as tracking it, however, and \cref{sup:f2} asks whether a model's answer changes correctly when the reference object changes.

\subsubsection{Instance Alignment of Box-Localization Errors}
\label{sup:f1-box-errors}

\textbf{Scope and rules.} A failed box may land on the wrong instance, or on the right instance with the wrong extent. To tell these apart, we align Sol's saved box predictions with annotated instances on 314 Natural door/window questions (23 buildings) and 113 BIM region questions (14 buildings), using the pixel contract and the IoU threshold of 0.60. Let $t$ be the maximum IoU with an accepted GT box, and $r_1,r_2$ the largest and second-largest IoUs with other annotated instances of the same kind. A \emph{wrong-instance} error has $r_1\geq0.60$, $r_1-\max(t,r_2)\geq0.10$, and a center outside all GT boxes. A \emph{near-target extent} error has its center inside a GT box and outside all other same-kind boxes, with $r_1<0.60$.

\begin{figure*}[t]
  \centering
  \includegraphics[width=\textwidth]{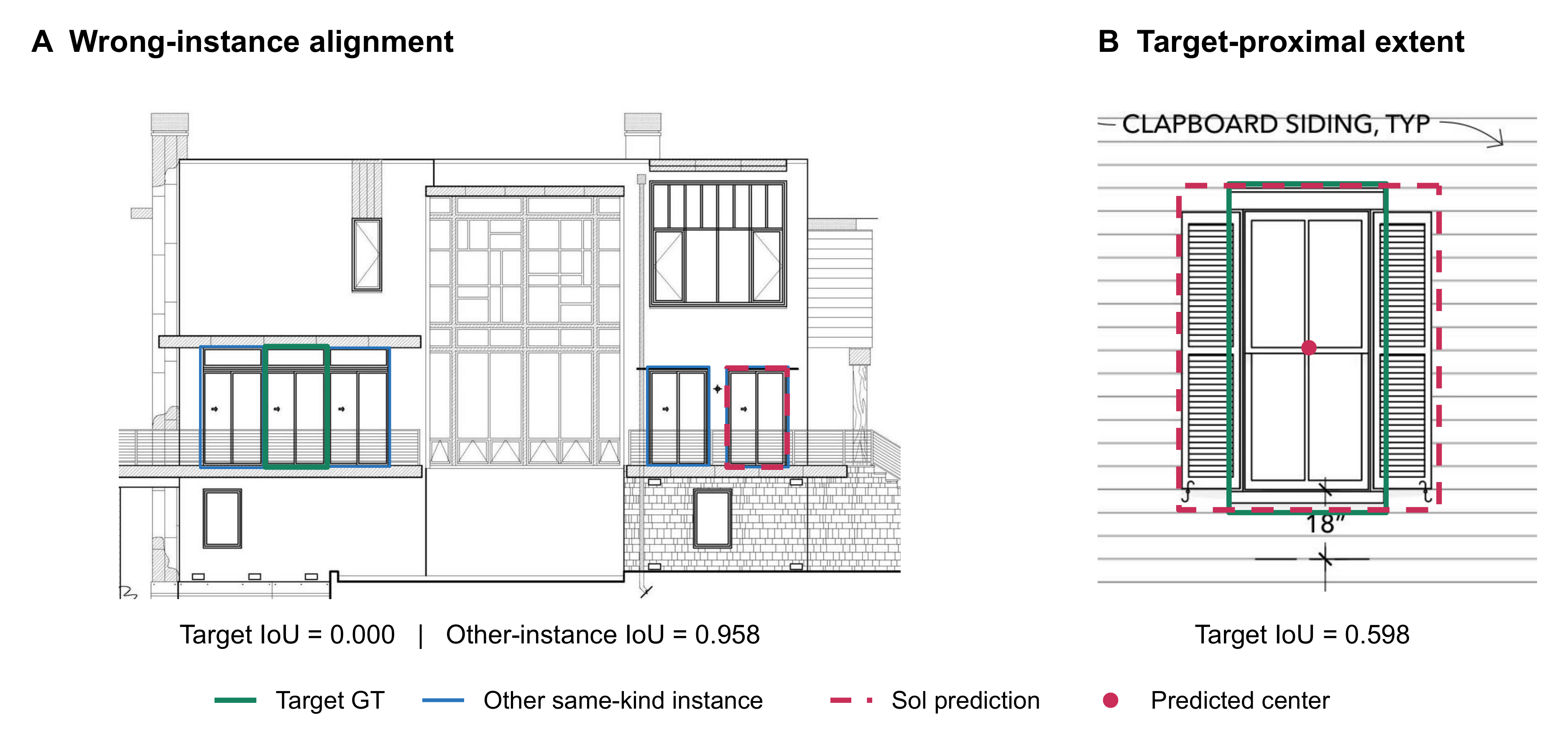}
  \caption{Two observed box-error patterns in Sol's saved predictions. Target drawings are cropped for display; scores use the original coordinates. A: the prediction overlaps another door (IoU 0.958) and misses the target (IoU 0). B: the center lies inside the target window, but the predicted extent is wider than the accepted box (IoU 0.598, below 0.60). Green denotes target GT, blue other same-kind instances, and dashed magenta the prediction; the dot in B marks its center.}
  \label{fig:box-instance-errors}
\end{figure*}

\textbf{Results.} Of Sol's 149 valid Natural errors, 24 are wrong-instance errors and 66 near-target extent errors; of its 52 valid BIM errors, the corresponding counts are 34 and six. \Cref{fig:box-instance-errors} shows one example of each. Box failures thus stem from both instance correspondence and region extent, a distinction that the center-point variant in \cref{sup:g3} builds on.

\subsection{Reference Switching and Repeated Querying}
\label{sup:f2}

\begin{figure*}[t]
  \centering
  \includegraphics[width=\textwidth]{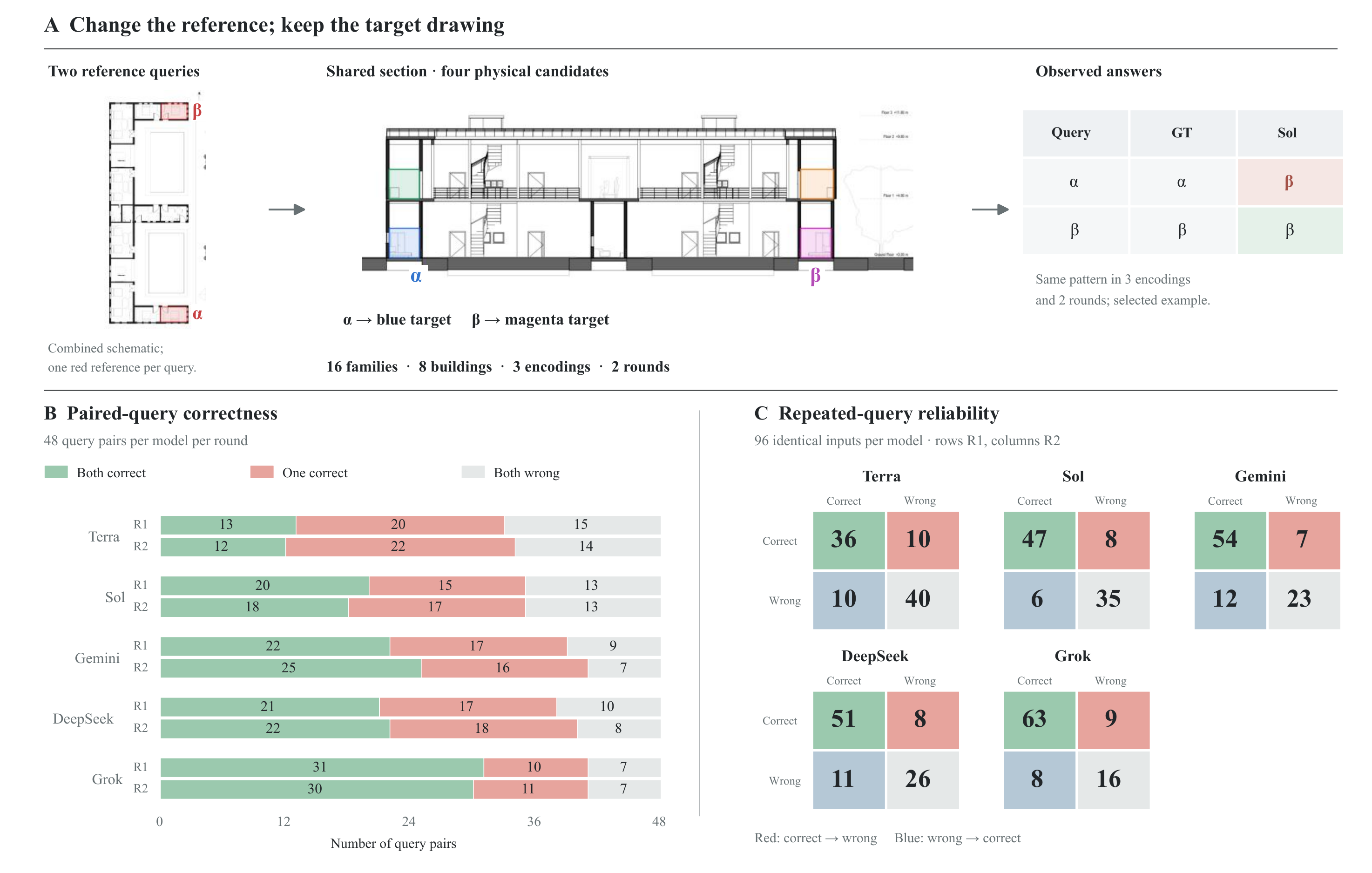}
  \caption{Reference switching and repeated querying. A combines two queries that differ only in switching the reference between $\alpha$/$\beta$; blue/purple mark the corresponding correct targets, and the table lists Sol's answers. B shows the per-round correctness counts over 48 query pairs for five models. C shows the two-round correctness transitions over 96 identical inputs per model, with rows for the first round and columns for the second; gray-red/gray-blue denote correct$\rightarrow$incorrect and incorrect$\rightarrow$correct, respectively.}
  \label{fig:S13}
  \label{sup:fig-s13-reference-switch}
\end{figure*}

Aggregate accuracy can hide unreliable correspondence. A model may answer one query in a scene correctly yet fail its counterpart, or may be right on a given input only some of the time. Unlike the Select--Locate pairing of \cref{sup:d3}, the answer format here stays fixed at candidate selection, and we vary either the reference object or nothing at all. The idea follows reliability measures from other settings: $\tau$-bench~\cite{yao2024taubench} counts a task as solved only if every repeated trial succeeds, and sample-consistency studies estimate confidence from the spread of answers over multiple generations~\cite{lyu2025calibrating}. We ground this perspective in physical entities. When the reference switches, the correct target must switch with it; when candidates are recolored or reordered, the chosen entity should stay the same; and when the input is repeated, a correct correspondence should persist.

\textbf{Two-round test of reference-target switching.} We build 16 question families from 8 BIM buildings. Within each family, the target drawing and candidates are fixed, and the reference mark switches between two instances $u$/$v$, so that the correct physical target switches too. Each family appears under three encodings (baseline, candidate recoloring, and option reordering), and each encoding is answered independently in two rounds with identical inputs. This gives each model 192 core responses, i.e., 48 query pairs per round, plus 24 simple-correspondence controls and 32 no-reference probes, for 248 responses. We test five models, Terra, Sol, Gemini, DeepSeek, and Grok, all at high reasoning effort, for 1,240 responses in total, 960 of them core.

Each response is mapped back to a physical instance through the letter/color mapping of its request, and each pair is classified as both correct, one correct, or both incorrect. Among the errors, we further distinguish selecting the same entity for both references, swapping $u$ and $v$, and picking a third distractor. Success on both queries measures whether the full pair is solved. \Cref{tab:S25} covers all 16 families. To weight buildings equally, we average over encodings and rounds within each family, then over families within each building, and finally over the 8 buildings.

\begin{table*}[t]
  \centering
  \footnotesize
  \setlength{\tabcolsep}{3pt}
  \caption{Two-round core results of reference switching for five models (16 families, 8 buildings, 48 query pairs per round). ``Same instance'' counts family $\times$ encoding conditions in which both references yield the same target instance within each round. The target can differ across rounds. The final column gives the number of families and buildings with this pattern.}
  \label{tab:S25}
  \begin{tabularx}{\textwidth}{@{}l*{5}{>{\centering\arraybackslash}X}@{}}
    \toprule
    Model & Round~1 both correct /48 & Round~2 both correct /48 & Two-round both correct, building-weighted & Same instance in both rounds, same encoding /48 & Families/buildings with this pattern \\
    \midrule
    Terra & 13 & 12 & 24.72\% & 7 & 5/3 \\
    Sol & 20 & 18 & 39.51\% & 11 & 5/2 \\
    Gemini & 22 & 25 & 46.81\% & 7 & 5/3 \\
    DeepSeek & 21 & 22 & 39.72\% & 10 & 5/3 \\
    Grok & 31 & 30 & 57.85\% & 6 & 4/3 \\
    \bottomrule
  \end{tabularx}
\end{table*}

\textbf{Outcomes for query pairs.} Complete pairs are far from guaranteed: models solve both queries in only 25--61 of their 96 core pairs. Some failures are systematic. In 5 of the 16 families, across 4 of the 8 buildings, at least one model selects the same member of the reference pair for both queries, in both rounds and under at least two encodings.

\textbf{Correctness changes under identical-input repetition.} Repeating the same input reveals a second kind of instability. Each model answers the same 96 core inputs twice, and per model 7--10 answers flip from correct to incorrect and 6--12 from incorrect to correct. Terra shows how this can hide behind a stable score: it answers 46/96 correctly in both rounds, yet 10 answers flip in each direction. \Cref{fig:S13}C gives the complete $2\times2$ matrices for strict correctness.

\textbf{Example.} \Cref{fig:S13}A shows a typical case. Two nearly mirrored bedrooms lie at opposite ends of one floor and look alike locally. Sol picks the same bedroom in all 12 responses across three encodings and two rounds, even though the answer letters and candidate colors change, and so answers exactly one query of each pair correctly.

Together, these results separate single-shot accuracy from instance-level reliability: a model may solve one query in a scene and fail its counterpart, and may change which questions it answers correctly while its aggregate score stays the same.

\subsection{Section-Line Direction and Position Decomposition}
\label{sup:f6}

To see how section lines fail, we decompose 1,080 saved answers from 12 models on the 90 directed section-line questions (50 Natural, 40 BIM). Because the aim is to diagnose content rather than format, we use the content criterion, which allows predefined deterministic format recovery; the strict accuracy of the main results additionally requires a compliant original output. For each scorable answer, we check the viewing direction and the two-end band test independently. When a line has the correct direction but fails geometry, and meets both checks at finite intersections, we record its maximum normal distance outside the band, and we call it parallel when its angle to the band axis in original-image coordinates is within $10^{-6}$ degrees. Correct-direction answers without finite intersections at both checks are counted separately.

\begin{table*}[t]
  \centering
  \footnotesize
  \setlength{\tabcolsep}{4pt}
  \caption{Mutually exclusive section-line states under the content criterion, which permits predefined deterministic format recovery. Columns distinguish correct viewing direction with accepted geometry, a parallel or nonparallel out-of-band line, or no finite two-end measurement. An unscorable or missing output remains in the fixed denominator.}
  \label{tab:S30}
  \begin{tabularx}{\textwidth}{llr*{7}{>{\raggedleft\arraybackslash}X}}
    \toprule
    & & & \multicolumn{4}{c}{Correct viewing direction} & \multicolumn{2}{c}{Wrong viewing direction} & \\
    \cmidrule(lr){4-7} \cmidrule(lr){8-9}
    Model & Source & $n$ & In band & Parallel outside & Nonparallel outside & No finite measure & In band & Geometry fails & Invalid or missing \\
    \midrule
    Terra & Natural & 50 & 8 & 11 & 0 & 0 & 1 & 30 & 0 \\
    Terra & BIM & 40 & 1 & 13 & 0 & 0 & 2 & 24 & 0 \\
    Sol & Natural & 50 & 20 & 17 & 0 & 0 & 3 & 10 & 0 \\
    Sol & BIM & 40 & 7 & 14 & 1 & 0 & 0 & 18 & 0 \\
    Gemini & Natural & 50 & 9 & 24 & 2 & 3 & 0 & 10 & 2 \\
    Gemini & BIM & 40 & 1 & 14 & 0 & 0 & 0 & 19 & 6 \\
    Grok-4.6 & Natural & 50 & 8 & 36 & 1 & 0 & 0 & 5 & 0 \\
    Grok-4.6 & BIM & 40 & 2 & 20 & 2 & 0 & 0 & 16 & 0 \\
    DeepSeek & Natural & 50 & 11 & 19 & 0 & 0 & 0 & 6 & 14 \\
    DeepSeek & BIM & 40 & 0 & 12 & 0 & 0 & 0 & 12 & 16 \\
    Qwen3.5-397B & Natural & 50 & 0 & 9 & 5 & 1 & 1 & 34 & 0 \\
    Qwen3.5-397B & BIM & 40 & 0 & 5 & 3 & 0 & 0 & 24 & 8 \\
    GLM-4.6V & Natural & 50 & 0 & 4 & 11 & 4 & 1 & 30 & 0 \\
    GLM-4.6V & BIM & 40 & 0 & 3 & 2 & 0 & 0 & 35 & 0 \\
    Kimi-K2.5 & Natural & 50 & 1 & 11 & 1 & 1 & 0 & 36 & 0 \\
    Kimi-K2.5 & BIM & 40 & 1 & 14 & 1 & 1 & 1 & 22 & 0 \\
    ERNIE & Natural & 50 & 1 & 10 & 3 & 5 & 0 & 30 & 1 \\
    ERNIE & BIM & 40 & 0 & 5 & 0 & 3 & 0 & 32 & 0 \\
    Mistral & Natural & 50 & 0 & 14 & 0 & 0 & 0 & 36 & 0 \\
    Mistral & BIM & 40 & 0 & 6 & 0 & 0 & 0 & 34 & 0 \\
    Llama & Natural & 50 & 1 & 8 & 0 & 2 & 3 & 35 & 1 \\
    Llama & BIM & 40 & 0 & 8 & 0 & 1 & 0 & 31 & 0 \\
    MiniMax & Natural & 50 & 0 & 4 & 4 & 1 & 1 & 30 & 10 \\
    MiniMax & BIM & 40 & 0 & 8 & 2 & 4 & 0 & 26 & 0 \\
    \bottomrule
  \end{tabularx}
\end{table*}

The most common outcome is failure of both the viewing-direction and geometry checks, with 292/600 Natural and 293/480 BIM answers. Even when the direction is right, the line usually lands far from the band. Of the 211 Natural and 142 BIM answers with the correct direction but failed geometry, 194 and 133 have measurable out-of-band distances, and most of these lines (167 and 122) are parallel to the band. They are misplaced rather than misdrawn: the median maximum distance outside the band is 9.12\% and 10.48\% of the image extent normal to the band, and only four measurable lines per source lie within 5 pixels of a boundary.

The mix differs by model. Sol has 17/50 Natural and 14/40 BIM answers with the correct direction and a parallel line outside the band, and Grok-4.6 has 36/50 and 20/40; all line outputs of both models are scorable after deterministic recovery. The BIM case in Fig.~5(a) of the main paper is typical. Sol and Gemini draw horizontal lines with the correct downward viewing direction but place them 192.49 and 112.49 pixels outside the band, toward the stair and entrance, where no cut could produce the reference section.

\subsection{Spatial Structure of Incorrect Choices}
\label{sup:f7}
\label{sup:f7-spatial-preferences}

\begin{figure*}[t]
  \centering
  \includegraphics[width=\textwidth]{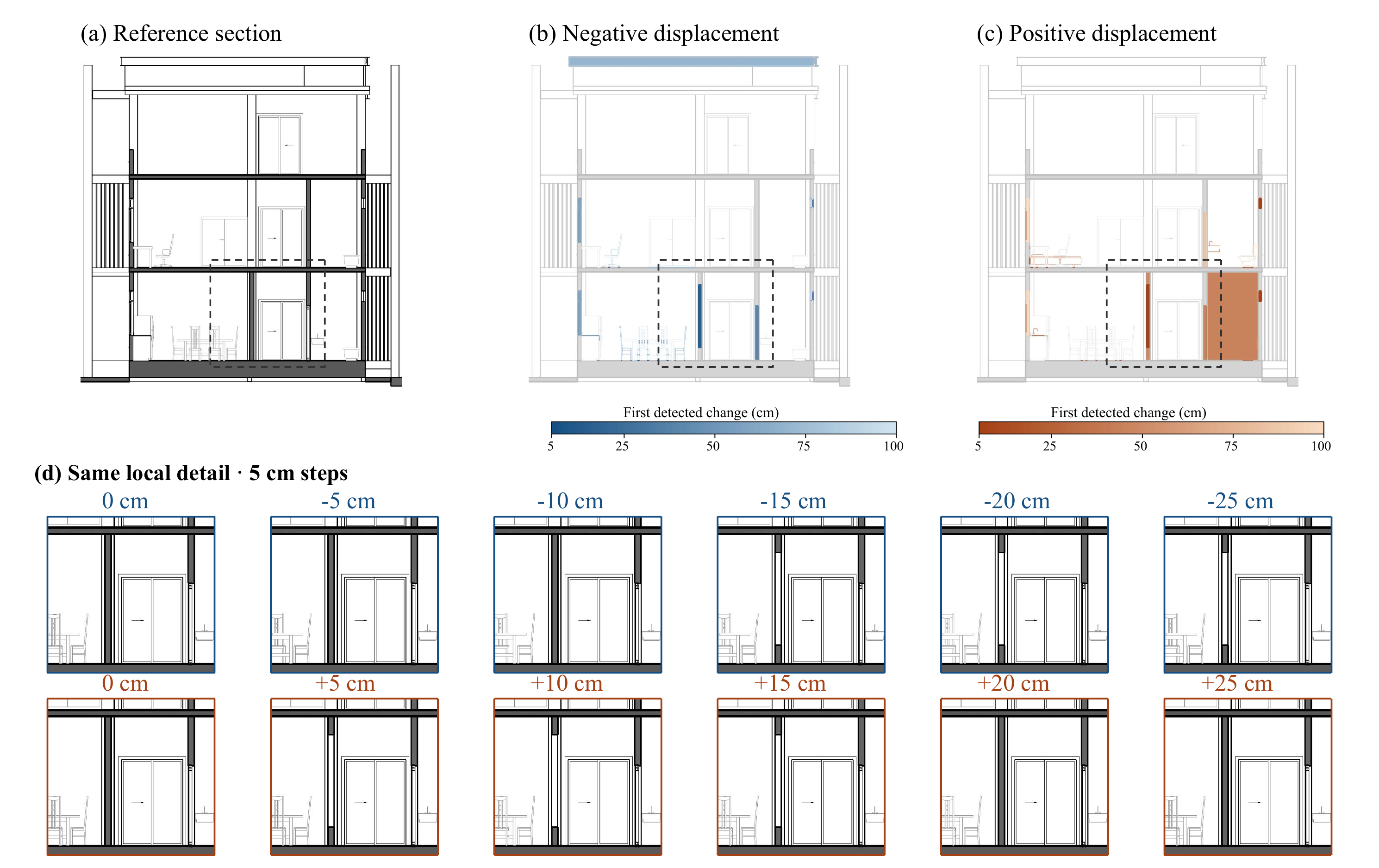}
  \caption{Local drawing differences caused by changes in section position. With the viewing conditions fixed, offsets of $\pm$1\,m are swept in 20\,cm steps (5\,cm for set F). (a)~Zero-offset reference; (b--c)~the sampled offsets at which a change is first detected in the negative (blue) and positive (orange) directions, where darker shades indicate smaller offsets and gray indicates no detected change; (d)~consecutive sections of the same local region from 0 to $\pm$25\,cm. While the overall outline remains similar, local fills and boundaries still change with the cut position.}
  \label{fig:S17}
  \label{sup:fig-s17-section-sensitivity}
\end{figure*}

\begin{table*}[t]
  \centering
  \footnotesize
  \setlength{\tabcolsep}{3pt}
  \caption{All 19 spatial measures for incorrect choices: correlation $r$ with excess incorrect votes, 95\% whole-building bootstrap interval, $q$ of the preference-weighted joint-choice test, and $q$ of the whole-building sign-flip test.}
  \label{tab:F7-1}
  \begin{tabularx}{\textwidth}{Lrrcrr}
    \toprule
    Measure & Questions & $r$ & 95\% building interval & Joint-choice $q$ & Whole-building $q$ \\
    \midrule
    Target surrounding layout, direct & 75 & $-0.040$ & $[-0.221, +0.095]$ & 0.5913 & 0.4603 \\
    Target mirrored surrounding layout & 75 & $+0.290$ & $[+0.156, +0.377]$ & 0.0046 & 0.0425 \\
    Target surrounding layout: best rotation/reflection & 75 & $+0.157$ & $[-0.061, +0.306]$ & 0.1434 & 0.3375 \\
    Horizontal mirror position & 75 & $+0.311$ & $[+0.147, +0.449]$ & 0.0084 & 0.0425 \\
    Vertical mirror position & 75 & $-0.254$ & $[-0.380, -0.120]$ & 0.1170 & 0.2830 \\
    Target 180$^\circ$ reversed position & 75 & $+0.090$ & $[-0.046, +0.257]$ & 0.3581 & 0.4045 \\
    Cross-view screen same-direction position & 75 & $+0.067$ & $[-0.037, +0.210]$ & 0.5913 & 0.5730 \\
    Cross-view horizontal screen flip & 75 & $-0.177$ & $[-0.363, -0.017]$ & 0.1014 & 0.3348 \\
    Cross-view vertical screen flip & 75 & $+0.070$ & $[-0.081, +0.251]$ & 0.1710 & 0.3637 \\
    Cross-view screen 180$^\circ$ reversal & 75 & $-0.160$ & $[-0.289, -0.037]$ & 0.5211 & 0.4099 \\
    Cross-view position: best rotation/reflection & 75 & $-0.010$ & $[-0.136, +0.157]$ & 0.4768 & 0.4099 \\
    Cross-view surrounding layout, direct & 75 & $+0.144$ & $[+0.049, +0.266]$ & 0.3581 & 0.3746 \\
    Cross-view mirrored surrounding layout & 75 & $+0.020$ & $[-0.174, +0.161]$ & 0.3581 & 0.4487 \\
    Cross-view layout: best rotation/reflection & 75 & $+0.017$ & $[-0.144, +0.161]$ & 0.3581 & 0.3838 \\
    Shared-axis forward position & 64 & $-0.181$ & $[-0.312, -0.055]$ & 0.3581 & 0.3838 \\
    Shared-axis reversed position & 64 & $+0.305$ & $[+0.172, +0.440]$ & 0.0084 & 0.0425 \\
    Horizontal screen flip $\times$ mirrored layout & 75 & $-0.157$ & $[-0.380, +0.022]$ & 0.1434 & 0.3746 \\
    Shared-axis reversal $\times$ mirrored layout & 64 & $+0.239$ & $[+0.101, +0.368]$ & 0.0407 & 0.0609 \\
    Local image mirror similarity & 75 & $+0.104$ & $[-0.006, +0.192]$ & 0.3581 & 0.4099 \\
    \bottomrule
  \end{tabularx}
\end{table*}

\textbf{Spatial distribution of incorrect choices.} We next ask whether wrong Select choices fall on particular candidates, which requires questions in which every candidate can be located precisely. Of the 882 Select questions, 534 bind four region candidates uniquely on the same target panel and verify against a same-source unmarked base drawing; 238 of these come from BIM. Restricting to component and space correspondence leaves 80, and excluding 5 space questions with duplicated incorrect candidate entities or geometry leaves 40 component and 35 space questions from 15 buildings. Each of these 75 questions has at least one valid incorrect response and passes verification of both drawings, the reference identity, the candidates, and the native-view binding.

From BIM entity IDs, the extent of visible entities in each drawing, and calibration on corresponding entities, we construct 19 measures of position, layout, and local appearance. Within each question, the three incorrect candidates are ranked by each measure and by excess incorrect votes (ties receive average ranks), and the centered ranks are pooled to compute the correlation. The 75 questions yield 225 incorrect candidates and 900 responses from 12 models. Shared-axis calibration needs at least six entities visible in both drawings, excluding the candidates, and checks the axis direction, anchor span, fitted direction, and residuals; 64 questions pass, and the other 11 are left out of the three shared-axis measures.

\textbf{Spatial measures.} Points are normalized by the extent of visible BIM entities in each view; the four candidate identities are excluded from layout counts and calibration anchors. \emph{Position} measures compare a candidate center with a transformed reference point (the correct target under horizontal reflection, vertical reflection, or $180^\circ$ rotation; or the source reference under identity, these transforms, or the best of the eight square symmetries) using $\exp(-d/0.15)$. \emph{Layout} measures compare category histograms of surrounding entity centers (12 categories, 4 radial rings, 8 sectors) by cosine similarity, under identity, the better axis reflection, or the best square symmetry. \emph{Shared-axis} measures fit target from source coordinates by Theil--Sen regression on at least six shared entities and score the candidate against the projected reference, either directly or reflected about the target midpoint. The two combined measures are geometric means of mirrored layout with horizontal screen reflection or shared-axis reversal. \emph{Local image mirror similarity} is the F1 overlap between the dilated ink mask of the correct candidate and the horizontally flipped mask of each incorrect candidate.

\textbf{Preference probabilities and correlation statistic.} Models also favor certain option positions and colors regardless of geometry, so we compare the observed votes with what these preferences alone would predict. For question $i$, let $W_i$ denote the set of three incorrect candidates and $E_i$ the set of models that give a valid incorrect choice. The probability $p_{mij}$ that model $m$ chooses incorrect candidate $j$ is a softmax over $W_i$ of one coefficient per option position plus one per color label, fitted with L2 regularization on the incorrect responses of that model in the other buildings. The observed incorrect votes are $v_{ij}=\sum_{m\in E_i}\mathbf{1}(C_{mi}=j)$, the preference-based expectation is $\mu_{ij}=\sum_{m\in E_i}p_{mij}$, and the excess incorrect votes are $e_{ij}=v_{ij}-\mu_{ij}$. For measure $k$, measure values and excess votes are ranked within the three incorrect candidates of each question and centered, giving $x_{ijk}$ and $y_{ij}$, and
\begin{equation}
  r_k=\frac{\sum_{i,j}x_{ijk}y_{ij}}{\sqrt{\left(\sum_{i,j}x_{ijk}^{2}\right)\left(\sum_{i,j}y_{ij}^{2}\right)}}.
\end{equation}

\textbf{Conditional test preserving the joint-choice structure.} Models often err together, so their choices are not independent, and the test preserves this structure. We fix $E_i$ for each question and the joint-choice grouping among models, and apply each of the six label permutations of the three incorrect candidates simultaneously to all incorrect choices. Models that originally chose the same candidate still choose a common candidate under every permutation. For permutation $\pi$, let $C_{mi}^{(\pi)}=\pi(C_{mi})$, with weight
\begin{equation}
  w_i(\pi)=\frac{\prod_{m\in E_i}p_{mi,C_{mi}^{(\pi)}}}{\sum_{\pi'}\prod_{m\in E_i}p_{mi,C_{mi}^{(\pi')}}}.
\end{equation}
After whole-group permutations are drawn with these weights, the votes, excess votes, ranks within each question, and correlation coefficients are recomputed; the preference probabilities and expected votes remain fixed. The test uses $B=50{,}000$ simulations; two-sided $p$ values with an add-one correction are BH-corrected across the 19 measures. Equal-probability permutations and independent option sampling serve as sensitivity checks.

\textbf{Building-level robustness.} Questions from one building can share layout regularities, so we also test at the building level. Each question's observed association contribution is reduced by its expected contribution under the weighted whole-group reference above, \ie, $s_{ik}=\sum_j x_{ijk}(y_{ij}-\mathbb{E}_{\pi}[y_{ij}^{(\pi)}])$, and the results are summed within buildings. Assuming that buildings are independent and that centered building scores are sign-symmetric under the null hypothesis, we enumerate all 32,768 sign combinations of the 15 buildings, compute two-sided $p$ values from the absolute total score, and apply BH correction across the 19 measures. The 95\% intervals of the correlation coefficients are obtained separately as the 2.5\% and 97.5\% quantiles of 2,000 whole-building bootstrap resamples. All four main positive correlations remain positive under leave-one-building-out. The $r$ ranges for horizontal mirror, shared-axis reversal, mirrored surrounding layout, and the combined measure are $[0.246, 0.341]$, $[0.274, 0.341]$, $[0.255, 0.323]$, and $[0.211, 0.273]$, respectively.

\textbf{Full results for the 19 measures.} Four positive correlations pass BH correction in the weighted joint-choice test, and all four describe mirror-like errors. Horizontal mirror position ($r=0.311$) and shared-axis reversed position ($r=0.305$) have $q=0.0084$, mirrored surrounding layout has $r=0.290$ and $q=0.0046$, and combining shared-axis reversal with mirrored layout yields $r=0.239$ and $q=0.0407$. The first three also reach $q=0.0425$ in the whole-building test, while the combined measure has $q=0.0609$. The equal-probability whole-group test and the independent-sampling reference yield 8 and 5 measures with $q<0.05$, respectively. \Cref{tab:F7-1} lists all 19 measures with their correlations, building bootstrap intervals, and $q$ values from both tests, each BH-corrected across the 19 measures.

Local appearance does not explain this preference. Local image mirror similarity shows only weak associations ($r=0.104$ over 75 questions and $0.067$ over the 64 shared-axis questions), with building intervals spanning zero.

\textbf{Example question in Fig.~4(a)--(b) of the main paper.} This example shows the position proxy at work. The reference is a room on the Floor 6 plan, and the four options in the target section are, in order, Magenta, Orange, Blue, and Green, with Orange correct. Eleven of the 12 models choose Magenta; only Kimi chooses Orange. The horizontal midpoint of the extent of visible entities in the target drawing is $x=1858.5$ and the GT center is $(2570.417, 330.0275)$, so the horizontal mirror point is $(1146.583, 330.0275)$, right next to the Magenta center at $x=1179.3395$. The figure shows the four candidates and the whole-building section together with the GT point, mirror point, and midline.

\subsection{Controlled Geometric Comparison of Cutting-Plane Translation}
\label{sup:f9}
\label{sup:f8}

Line localization rests on the premise that the cut position matters even when the overall section looks the same. A section combines the cross-sectional profiles of intersected elements with visible projections beyond the cut~\cite{choi2018isarc}, so moving the cutting plane can change what is drawn while the viewing direction stays fixed. We measure how quickly this happens.

\textbf{Sweep setup.} On six sections from three BIM buildings, we fix the viewing direction, framing, scale, and far clipping, and sweep $\pm$1\,m along the normal. The six sets use a 20\,cm step, and set F is additionally swept at a finer 5\,cm step, for 41 positions in total. The heat maps in \cref{fig:S17} show the offset at which a change relative to the zero-offset reference is first detected, using a grayscale threshold of 24/255 and a 1-pixel neighborhood tolerance. In the fine sweep, local fills change at positive offsets of 5--15\,cm and recover at 20\,cm. Local section evidence can therefore respond to cut shifts of a few centimeters while the overall outline stays the same, and it is this evidence that confines a compatible cut to the narrow bands of \cref{sup:b1-line-band}.

\begin{table*}[t]
\centering
\footnotesize
\setlength{\tabcolsep}{3pt}
\caption{Curriculum SFT recipe. Exposures count repetitions and replay.}
\label{tab:S31a}
\begin{tabularx}{\textwidth}{LLL}
\toprule
Training setting & Stage 1: Judge/Select & Stage 2: boxes and lines + closed-form replay \\
\midrule
Training-stream composition (exposures) & 10,831 closed-form questions once each, plus 1 to fill the batch & Region and line questions 3,431$\times$2 + closed-form 762 \\
Total exposures & 10,832 & 7,624 \\
Answer format & Option letter & \texttt{bbox\_2d} box, or two endpoints with a viewing direction, in 0--1000 coordinates \\
Updated parameters & Language LoRA & Language LoRA + vision merger \\
Learning rate (LoRA / merger) & $5\times10^{-5}$ / frozen & $5\times10^{-5}$ / $10^{-5}$ \\
Global batch / optimizer update steps & 16 / 677 & 4 / 1,906 \\
Checkpoint used & Step 677 & Step 953, selected on the development buildings \\
\bottomrule
\end{tabularx}
\end{table*}

\begin{table*}[t]
\centering
\footnotesize
\setlength{\tabcolsep}{3pt}
\caption{Geometric contracts of the main benchmark and the center-point variant.}
\label{tab:S31}
\begin{tabularx}{\textwidth}{lLL}
\toprule
Localization requirement & Main benchmark & Center-point variant \\
\midrule
Region correspondence & Output a four-corner box in pixel coordinates, scored by the region-geometry rules of \cref{sup:c2} & Output an integer center point in 0--1000; it must fall inside the original GT box with center error / GT-box diagonal $\leq 0.25$ \\
Section line & Correct viewing direction, and the supporting line lies within the accepted positional band (boundaries inclusive) over the GT evaluation span & Correct viewing direction, and mean normalized endpoint error $\leq 0.055$, with swapped endpoint order allowed \\
\bottomrule
\end{tabularx}
\end{table*}

\section{Training Method and Results}
\label{sup:g}

This section details the fine-tuning study of Sec.~6 of the main paper. We describe the data split, curriculum, and scoring (\cref{sup:g1}), report results on BIM and Natural drawings (\cref{sup:g2}), and then examine a variant that supervises center points instead of boxes (\cref{sup:g3}).

\subsection{Data, Curriculum, and Scoring}
\label{sup:g1}

All training data come from BIM. The 23 BIM buildings are split by building into 15 training, four development, and four test buildings, and the 14,262 training questions comprise 6,506 Judge, 4,325 Select, 3,325 region, and 106 line questions.

\paragraph{Curriculum SFT.}
We fine-tune Qwen3.5-9B in two stages, ordered by the demand of the answer format (\cref{tab:S31a}). The model first learns all 10,831 Judge/Select questions, and then the 3,431 region and line questions, each seen twice and mixed with 762 replayed closed-form questions. Region answers are boxes written as \texttt{bbox\_2d=[xmin,ymin,xmax,ymax]}, and line answers are two endpoints with a viewing direction; all coordinates are normalized to 0--1000 by image width and height, and the prompt states this rule together with the target image index and the JSON answer format. Training uses plain cross-entropy on the answer JSON and EOS, with no geometric reward and no new localization head. We train rank-16 LoRA adapters on the language side while freezing the original language weights and the vision encoder; the second stage additionally updates the vision merger. Of its 1,906 optimization steps, we keep the checkpoint at step 953, selected on the development buildings.

Both stages use LoRA rank 16, alpha 32, dropout 0.05, AdamW, weight decay 0.01, and gradient clipping at 1.0, and image and prompt tokens are excluded from the loss.

\textbf{Evaluation and scoring.} On BIM, box and line localization use the 2,004 free-form questions of the four test buildings, and Judge/Select use the 441 Judge and 447 Select questions of the BIM benchmark, which are excluded from training; on Natural, all formats use the 1,672 benchmark questions. Decoding is greedy, with thinking disabled, at most 256 generated tokens, and a per-image processing cap of 2,007,040 pixels. A predicted box passes at IoU $\geq 0.60$ with the GT box, as in the main benchmark. A BIM line passes when its viewing direction is correct and its mean normalized endpoint error is at most 0.055, with either endpoint order allowed. Natural questions use the scoring of the main benchmark, and invalid or truncated responses count as incorrect.

\subsection{Results on BIM and Natural Drawings}
\label{sup:g2}

\begin{table}[t]
\centering
\footnotesize
\setlength{\tabcolsep}{3pt}
\caption{Original and fine-tuned Qwen3.5-9B. BIM Box/Line use the four held-out buildings; Natural is not used for training. Judge/Select: accuracy (\%) with chance-corrected score in parentheses; Box/Line: accuracy (\%).}
\label{tab:S32}
\begin{tabular}{@{}lrrrr@{}}
\toprule
Model & Judge & Select & Box & Line \\
\midrule
\multicolumn{5}{@{}l}{\textit{BIM (441 / 447 / 1,976 / 28)}} \\
Original & 56.7 (13.4) & 32.7 (10.2) & 0.00 & 0.00 \\
Fine-tuned & 62.1 (24.3) & 45.4 (27.2) & 25.30 & 7.14 \\
\midrule
\multicolumn{5}{@{}l}{\textit{Natural (794 / 435 / 393 / 50)}} \\
Original & 61.0 (21.9) & 35.2 (13.6) & 0.25 & 2.00 \\
Fine-tuned & 64.6 (29.2) & 50.1 (33.5) & 30.79 & 2.00 \\
\bottomrule
\end{tabular}
\end{table}

Fine-tuning turns a model that localizes nothing into one that localizes a quarter of the held-out boxes. It succeeds on 500 of the 1,976 held-out boxes (25.30\%) and 2 of the 28 lines (7.14\%), or 502 of all 2,004 free-form questions (25.05\%), whereas the original model localizes none (\cref{tab:S32}). Closed-form answers improve as well: BIM Judge rises from 250/441 (56.69\%) to 274/441 (62.13\%) and Select from 146/447 (32.66\%) to 203/447 (45.41\%), gains of 5.4 and 12.8 percentage points in raw accuracy, or 10.9 and 17.0 chance-corrected points.

\textbf{Transfer to Natural drawings.} The gains carry over to real drawing sets that play no part in fine-tuning or checkpoint selection. On Natural, box accuracy rises from 1/393 (0.25\%) to 121/393 (30.79\%), Select from 153/435 (35.17\%) to 218/435 (50.11\%), and Judge from 484/794 (60.96\%) to 513/794 (64.61\%), while lines stay at 1/50. Pooling boxes and lines, Natural localization rises from 2/443 (0.45\%) to 122/443 (27.54\%).

\subsection{Center-point Variant and Transfer to Natural Drawings}
\label{sup:g3}

The box results leave open which part of the task the model has learned. A box answer must find the corresponding instance and also recover its extent, and Sec.~4.3 of the main paper shows that box errors involve both. The center-point variant isolates the first part: it keeps the images, target identity, and reference marks, but asks only for the target's center point. With this output, the fine-tuned 9B model locates 31.8\% of held-out BIM targets and 46.1\% of Natural targets, compared with box accuracies of 25.3\% and 30.8\% under the box curriculum (\cref{tab:S32}). BIM supervision therefore teaches the model to find the corresponding instance across views, and much of the remaining box error lies in the extent. The same curriculum also improves a 27B model (\cref{tab:S32b}).

\textbf{Adapted outputs.} Each point question asks for the center of the originally annotated box. Adding 1,533 point questions from the training buildings gives 4,858 point supervision questions in total. \Cref{tab:S31} compares the adapted scoring with that of the main benchmark; points and lines must also name the correct target drawing and use strict JSON with integer coordinates.

\textbf{Curriculum.} The variant reuses the closed-form stage. It then learns the 3,325 original and 1,533 additional point questions (original points seen twice) with closed-form replay while updating the vision merger, and finally adds the 106 line questions with point and closed-form replay, keeping the merger fixed. All other settings follow \cref{sup:g1}.

\textbf{Natural transfer set.} We evaluate on all 1,672 Natural benchmark questions (23 buildings, 1,470 images) with the same adapted prompts and scoring. None is used for training or selection, and the 15 training buildings (9,569 images) share no building, question ID, or encoded-image hash with them.

\textbf{Held-out BIM buildings.} On the held-out buildings, point accuracy rises from 2.94\% to 31.78\% (1,976 questions) and line accuracy from 0/28 to 3/28; pooled, localization rises from 58/2,004 (2.89\%) to 631/2,004 (31.49\%). Point accuracy rises in each of the four test buildings, by 24.5 to 44.1 percentage points, and the equal-building mean goes from 4.07\% to 35.70\%.

\textbf{Closed-form answers and model scale.} Closed-form answers improve too. On the 3,377 Judge and 2,487 Select questions of the four test buildings, the fine-tuned 9B model raises Judge from 52.09\% to 65.21\% and Select from 49.34\% to 70.00\% (\cref{tab:S32b}). Applying the same center-point curriculum to Qwen3.8-27B improves every column as well, so the effect of BIM supervision holds at a larger scale.

\begin{table}[t]
\centering
\footnotesize
\setlength{\tabcolsep}{3pt}
\caption{Center-point variant at two model scales, original $\rightarrow$ fine-tuned (accuracy, \%). BIM columns use the four held-out buildings; Open pools points and lines. Natural is not used for training.}
\label{tab:S32b}
\begin{tabular}{@{}lrrrr@{}}
\toprule
 & \multicolumn{3}{c}{BIM held-out} & Natural \\
\cmidrule(lr){2-4}\cmidrule(l){5-5}
Model & Judge /3,377 & Select /2,487 & Open /2,004 & Open /443 \\
\midrule
9B & 52.1$\rightarrow$65.2 & 49.3$\rightarrow$70.0 & 2.9$\rightarrow$31.5 & 3.8$\rightarrow$41.8 \\
27B & 56.6$\rightarrow$66.7 & 51.6$\rightarrow$72.7 & 14.3$\rightarrow$33.9 & 27.1$\rightarrow$44.2 \\
\bottomrule
\end{tabular}
\end{table}

\textbf{Transfer to Natural drawings.} Again without Natural training, point accuracy rises from 17/393 (4.33\%) to 181/393 (46.06\%), with gains in all 23 Natural buildings and an equal-building macro accuracy that rises from 4.24\% to 44.06\%. Predicted points also land closer: on the 376 points answered validly by both checkpoints, the median center error falls from 1.296 to 0.390 box diagonals. Lines improve only from 0/50 to 4/50.

\textbf{Tighter tolerances.}
\label{sup:g4}
\label{sup:g3-point-pairing}
\begin{table}[t]
\centering
\footnotesize
\setlength{\tabcolsep}{4pt}
\caption{Center-point variant on held-out BIM (/2,004, accuracy \%) under tighter point-center / line-endpoint tolerances. The first row drops the center-distance requirement.}
\label{tab:S34}
\begin{tabular}{@{}lrr@{}}
\toprule
Tolerance & Original & Fine-tuned \\
\midrule
Inside box / 0.055 & 5.64 & 33.68 \\
0.25 / 0.055 (primary) & 2.89 & 31.49 \\
0.10 / 0.020 & 0.90 & 25.85 \\
0.05 / 0.010 & 0.30 & 21.91 \\
0.02 / 0.005 & 0.00 & 11.83 \\
0.01 / 0.005 & 0.00 & 6.24 \\
\bottomrule
\end{tabular}
\end{table}

\Cref{tab:S34} rescores the same answers under progressively tighter tolerances. The fine-tuned model keeps a large margin throughout; even at 0.05/0.010, accuracy rises from 0.30\% to 21.91\%. Center errors on BIM shrink as well: on the points answered validly by both checkpoints, the median falls from 2.468 to 0.770 box diagonals.

\end{document}